\documentclass[lettersize,journal,twoside]{IEEEtran}
\usepackage{amsmath,amsfonts}
\pdfoutput=1
\usepackage{algorithmic}
\usepackage[ruled,vlined]{algorithm2e}
\usepackage{array}
\usepackage[caption=false,font=normalsize,labelfont=sf,textfont=sf]{subfig}
\usepackage{textcomp}
\usepackage{stfloats}
\usepackage{url}
\usepackage{verbatim}
\usepackage{graphicx}
\def\BibTeX{{\rm B\kern-.05em{\sc i\kern-.025em b}\kern-.08em
    T\kern-.1667em\lower.7ex\hbox{E}\kern-.125emX}}
\usepackage{balance}
\usepackage{amssymb}
\usepackage{verbatim}
\usepackage{multirow}
\usepackage{booktabs}
\usepackage{bbding}
\usepackage{pifont}
\usepackage{makecell}
\usepackage{soul, color}
\usepackage{setspace}
\usepackage{colortbl}
\usepackage{amsmath}
\usepackage[colorlinks,
            linkcolor=blue,
            anchorcolor=blue,
            citecolor=blue]{hyperref}
\usepackage{orcidlink}
\hypersetup{hidelinks,colorlinks=true,allcolors=black,pdfstartview=Fit,breaklinks=true}

\usepackage[table,dvipsnames]{xcolor}
\definecolor{lightgray}{rgb}{0.92, 0.92, 0.92}

\usepackage{soulutf8}
\soulregister\cite7 
\soulregister\ref7
\soulregister\eqref7
\soulregister{\citep}7 
\soulregister{\citet}7 
\usepackage{cite}

\begin{document}
\title{Unseen from Seen: Rewriting Observation-Instruction Using Foundation Models for Augmenting Vision-Language Navigation}
\author{Ziming Wei\orcidlink{0009-0008-5350-7805}, Bingqian Lin\orcidlink{0000-0002-8763-9530}, Yunshuang Nie\orcidlink{0009-0000-9717-2308}, Jiaqi Chen\orcidlink{0009-0002-5126-3460}, Shikui Ma\orcidlink{0000-0002-8340-9762}, Hang Xu\orcidlink{0000-0003-3645-8972}, and Xiaodan Liang\orcidlink{0000-0003-3213-3062}
\thanks{Received 10 September 2024; revised 23 March 2025, 11 July 2025, and 11 August 2025; accepted 15 October 2025. This work is supported by National Key Research and Development Program of China (2024YFE0203100), National Natural Science Foundation of China (NSFC) under Grants No.62476293, National Postdoctoral Program for Innovative Talents under Grant Number BX20250379, China Postdoctoral Science Foundation under Grant Number 2025M771521, The Major Key Project of PCL (No. PCL2024A04, No. PCL2025A12-2), Guangdong S\&T Programme under Grant 2024B0101010003, and General Embodied AI Center of Sun Yat-sen University. \it {(Ziming Wei and Bingqian Lin contributed equally to this work.) (Corresponding author: Xiaodan Liang.)}}
\thanks{Ziming Wei and Yunshuang Nie are with Shenzhen Campus of Sun Yat-sen University, Shenzhen 518107, China. (e-mail: weizm3@mail2.sysu.edu.cn; nieysh@mail2.sysu.edu.cn).}
\thanks{Bingqian Lin is with Shanghai Jiao Tong University, Shanghai 200240, China. (e-mail: linbq666@sjtu.edu.cn).}
\thanks{Jiaqi Chen is with The University of Hong Kong, Hong Kong, China. (e-mail: jqchen@cs.hku.hk).}
\thanks{Shikui Ma is with Hunan Artificial Intelligence and Robotics Institute Company Ltd., Changsha 410100, China. (e-mail: zn005588@hnlens.com).}
\thanks{Hang Xu is with Huawei Noah's Ark Lab, Shanghai 201206, China. (e-mail: chromexbjxh@gmail.com).}
\thanks{Xiaodan Liang is with Shenzhen Campus of Sun Yat-sen University, Shenzhen 518107, China, and also with Peng Cheng Laboratory, Shenzhen 518066, China. (e-mail: liangxd9@mail.sysu.edu.cn).}
\thanks{This article has supplementary downloadable material available at https://doi.org/10.1109/TNNLS.2025.3624691, provided by the authors.}
\thanks{Code is available at \href{https://github.com/SaDil13/VLN-RAM}{https://github.com/SaDil13/VLN-RAM}.}
\thanks{Digital Object Identifier 10.1109/TNNLS.2025.3624691}
}

\markboth{IEEE TRANSACTIONS ON NEURAL NETWORKS AND LEARNING SYSTEMS}%
{WEI \textit{et al.}: UNSEEN FROM SEEN: REWRITING OBSERVATION-INSTRUCTION USING FOUNDATION MODELS}
\IEEEaftertitletext{\vspace{-4.0em}}

\maketitle


\begin{abstract}
Data scarcity is a long-standing challenge in the Vision-Language Navigation (VLN) field, which extremely hinders the generalization of agents to unseen environments. Previous works primarily rely on additional simulator data or web-collected images/videos to improve the generalization. However, the simulator environments still face limited diversity, and the web-collected data often requires extensive labor to remove the noise. 
In this paper, we propose a \textbf{R}ewriting-driven \textbf{A}ug\textbf{M}entation (RAM) paradigm for VLN, which directly creates the unseen observation-instruction pairs via rewriting human-annotated training data. Benefiting from our rewriting mechanism, new observation-instruction pairs can be obtained in both simulator-free and labor-saving
manners to promote generalization. Specifically, we first introduce Object-Enriched Observation Rewriting, where we combine Vision-Language Models (VLMs) and Large Language Models (LLMs) to derive rewritten object-enriched scene descriptions, enabling observation synthesis with diverse objects and spatial layouts via Text-to-Image Generation Models (T2IMs). Then, we propose Observation-Contrast Instruction Rewriting,
which generates observation-aligned rewritten instructions by requiring LLMs to reason the difference between original and new observations.
We further develop a mixing-then-focusing training strategy with a random observation cropping scheme, effectively enhancing data distribution diversity while suppressing augmentation data noise during training. 
Experiments on both the discrete environments (R2R, REVERIE, and R4R datasets) and continuous environments (R2R-CE dataset) show the superior performance and impressive generalization ability of our method.
\end{abstract}


\section*{Nomenclature}

\begin{IEEEdescription}[\IEEEusemathlabelsep]
\item[$I$] Instruction.
\item[$O_{t}$] Observation.
\item[$C_{t}$] Scene description.
\item[$C^{r}_{t}$] Rewritten object-enriched scene description.
\item[$O_{t}^{r}$] Rewritten observation.
\item[$G_{t}$] Ground-truth action (observation).
\item[$U_{t}$] Grounded landmark for the ground-truth action (observation).
\item[$C^{'}_{t}$] Scene description of the rewritten observation.
\item[$I^{r}$] Rewritten instruction.
\item[$P_{c}$] Scene description rewriting prompt.
\item[$P_{i}$] Instruction rewriting prompt.

\end{IEEEdescription}

\section{Introduction}
\IEEEPARstart{V}{ision}-Language Navigation (VLN)
\cite{anderson2018vision},\cite{qi2020reverie},\cite{ku2020room},\cite{thomason2019vision},
as a critical research path to achieve embodied intelligence, defines the task that requires an embodied agent to navigate through complex 3D environments following natural language instructions. 
To enable VLN agents to learn the observation-instruction alignment knowledge and navigation skills, researchers have provided elaborately annotated trajectory-instruction data from the simulator environments~\cite{chang2017matterport3d} for navigation training.
Nevertheless, the limited amount of high-quality manually annotated VLN data, i.e., the data scarcity issue, greatly harms the generalization of existing agents to unseen environments with various possible spatial layouts and co-occurrence objects. 

To address the data scarcity problem, prior studies have investigated numerous data augmentation strategies, which can be roughly divided into two categories: simulator-based and web-based. Simulator-based methods
\cite{tan2019learning},\cite{fried2018speaker},\cite{liu2021vision},\cite{chen2022learning},\cite{wang2023scaling},\cite{fu2020counterfactual} 
rely on the original Matterport3D~\cite{chang2017matterport3d} environment or additional simulators like HM3D~\cite{ramakrishnan2021habitat} and Gibson~\cite{xia2018gibson} to perform augmentation trajectory sampling and use the Speaker~\cite{fried2018speaker} model for instruction generation. In contrast, web-based approaches
\cite{Guhur2021AirbertIP}, \cite{lin2023learning}, collect room images or room-tour videos from the web for trajectory augmentation, and adopt the template-based strategy to generate instructions. 
Despite the promising performance improvement, these approaches still have the following limitations.
For trajectory generation, the simulator data is always constrained in specific environments and hard to extend to various scenarios. 
On the other hand, web images/videos usually contain extensive noise and demand labor-intensive data cleaning. For instruction augmentation, Speaker-generated instructions are usually uninformative while template-based instructions may have limited flexibility.

\definecolor{moti_red}{RGB}{192,0,0}
\definecolor{moti_blue}{RGB}{0,153,255}

\begin{figure*}[t]
\begin{centering}
\includegraphics[width=\linewidth]{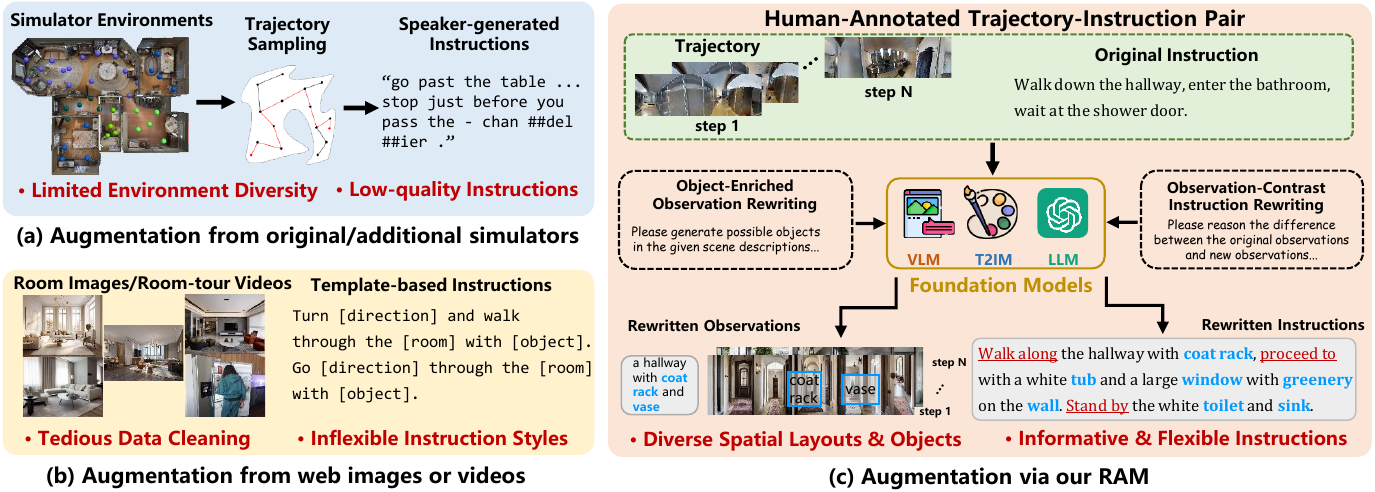}
\par\end{centering}
\caption{Comparison of our RAM with typical VLN data augmentation approaches: (a) Augmentation from original/additional simulators and (b) Augmentation from web images or videos. Rather than these methods that may be limited in specific simulator environments or struggle with tedious data cleaning, our RAM (c) empowers simulator-free and labor-saving
data augmentation by rewriting human-annotated data to generate unseen observation-instruction pairs.
The generated scene objects and the newly introduced objects mentioned in the rewritten instructions are denoted by \textcolor{moti_blue}{\textbf{blue}}  boxes/fonts. New actional representations in the rewritten instructions are denoted by \textcolor{moti_red}{\underline{red underlined}} fonts. 
The mixing-then-focusing training strategy is omitted in this figure.}
\label{fig:motivation}
\end{figure*}

In this paper, we propose a new data augmentation framework, dubbed \textbf{R}ewriting-driven \textbf{A}ug\textbf{M}entation (RAM), which generates the unseen observation-instruction pairs via rewriting original human-annotated data.
Our RAM enables observation-instruction rewriting in both {\it simulator-free} and {\it labor-saving} 
manners, powered by an elegant combination for various foundation models. As shown in Fig.~\ref{fig:motivation}, our RAM encompasses two core modules. Firstly, we introduce \textbf{Object-Enriched Observation Rewriting}, which synthesizes new observations with different spatial layouts and objects out of the simulator while requiring no labor-intensive 
postprocessing procedures. 
To achieve this, we combine Vision-Language Models (VLMs) and Large Language Models (LLMs) to obtain object-enriched rewritten observation descriptions. Then we feed them to Text-to-Image Generation Models (T2IMs) to perform new observation generation.
Secondly, we conduct \textbf{Observation-Contrast Instruction Rewriting}, where we ask LLMs to generate observation-aligned instructions by reasoning the difference between original and new observation descriptions obtained via VLMs. 
We also require the LLMs to change the representations of actional descriptions and syntax architecture for generating new instructions. Therefore, our rewritten instructions are more informative with flexible representations. 
To strengthen the diversity of data distribution while suppressing the negative impact of rewritten data (e.g., the inherent generation noise brought by T2IMs) for improving training, we further develop a \textbf{mixing-then-focusing training scheme} accompanied by a random observation cropping strategy.
As a result, our rewritten data can bring significant performance gains with only a small scale. 

Experimental results on R2R~\cite{anderson2018vision}, REVERIE~\cite{qi2020reverie}, and R4R~\cite{jain2019stay} show the superior performance of RAM over existing methods especially in unseen scenarios, demonstrating the effectiveness of our proposed rewriting-driven augmentation paradigm in improving generalization.
Through only small-scale data augmentation, our RAM achieves comparable or even better performance to a recent State-Of-The-Art (SOTA) method~\cite{wang2023scaling} that relies on large-scale data from additional simulators on R2R and REVERIE. The transfer of our RAM to the R2R-CE~\cite{krantz2020beyond} dataset further reveals the promising generalization ability of the proposed method. 

To summarize, the main contributions of this work are follows. 
\vspace{-2em}
\begin{itemize}
\setlength{\itemsep}{0pt}
\setlength{\parsep}{0pt}
\setlength{\parskip}{0pt}
  \item[1)]
  We propose RAM,  a novel VLN data augmentation paradigm that rewrites the human-annotated data to generate unseen observation-instruction pairs. The proposed framework integrates a series of foundation models to perform rewriting without any reliance on the simulator environments or web-collected data.
  \item [2)]
  We provide a detailed evaluation of different data fusion schemes for training.
  We also introduce a two-stage mixed training strategy with a random observation cropping scheme, which sufficiently activates the advantage of our rewritten data while simultaneously mitigating its inherent noise to boost training.
  \item[3)] RAM shows impressive generalization ability on multiple VLN datasets. It achieves comparable or even better performance to a recent SOTA data augmentation method by introducing $\sim$100 times less augmentation data.
  \end{itemize}

The remainder of this paper is organized as follows. Section~\ref{related work} provides a brief review of the related work. Section~\ref{method} describes the problem setup of the VLN task and then introduces our proposed method. Section~\ref{experiments} presents the experimental setup and results. Section~\ref{conclusion} concludes the paper and gives some outlook for future work.

\section{Related Work}
\label{related work}

\subsection{Vision-Language Navigation}
Vision-Language Navigation (VLN) is a challenging task that asks an agent to follow human instructions to navigate to the target position. As a core task in the Embodied AI area, VLN has drawn increasing research interest in recent years.
Early methods use a sequence-to-sequence architecture\cite{anderson2018vision} to build VLN models with cross-modal alignment modules
\cite{ma2019self}, \cite{Qi2020ObjectandActionAM} and 
effective training mechanisms\cite{wang2019reinforced},\cite{zhu2020vision},\cite{Wang2020SoftER},\cite{9615119} such as reinforcement learning\cite{wang2019reinforced,tan2019learning}, contrastive learning\cite{10120966}, \cite{9954217}, \cite{liang2022contrastive}, adversarial learning\cite{lin2021adversarial,fu2020counterfactual}, {\it etc}.
For example, RCM~\cite{wang2019reinforced}
introduces a reinforced cross-modal matching approach to enforce the cross-modal grounding both locally and globally via reinforcement learning. 
AuxRN\cite{zhu2020vision} proposes multiple self-supervised auxiliary learning tasks to explore the rich information in the VLN environment.
DISH~\cite{10530862} introduces a hierarchical reinforcement learning method for VLN, decomposing tasks into subgoals via a manager-worker framework to alleviate the reward sparsity. GL-VLN~\cite{9813501} proposes a global normalization strategy with two score functions to rerank beam-search trajectories in VLN, addressing trajectory-level score bias. VPG~\cite{9615119} proposes a meta-learning-based visual perception generalization strategy with MAML and feature-wise affine transformation, achieving effective skill transfer in VLN tasks.

Inspired by the great success of vision-language pretraining
\cite{li2020unicoder},\cite{li2020oscar},\cite{chen2020uniter},\cite{lu2019vilbert},\cite{Li2019VisualBERTAS}, recent approaches use transformer-based architecture and devise domain-specific proxy tasks to boost the performance of VLN models
\cite{hao2020towards},\cite{hong2021vln},\cite{Chen2021HistoryAM},\cite{Chen2022ThinkGA},\cite{an2023bevbert},\cite{Zhao2022TargetDrivenST},\cite{Qi2021TheRT},\cite{tan2024self},\cite{10495141},\cite{10359152},\cite{11071855},\cite{10414007},\cite{10531078}.
PREVALENT~\cite{hao2020towards} constructs large-scale image-text-action triplets to pretrain the agent. 
VLN$\circlearrowright$BERT~\cite{hong2021vln} enables the agent to recognize time-dependent input by introducing a recurrent function. 
HAMT~\cite{Chen2021HistoryAM} constructs a history-aware multimodal transformer for long-horizon navigation history encoding. 
However, most existing methods rely on domain-specific VLN data obtained from limited simulator environments, and therefore cannot generalize well to various unseen scenarios. 

Some recent works have attempted to unleash the rich world knowledge from the foundation models, such as Large Language Models (LLMs)~\cite{openai},\cite{OpenAI_2023} and Vision Language Models (VLMs)~\cite{radford2021learning}, \cite{dai2023instructblip}, to improve the generalization ability of the VLN agent
\cite{zhou2023navgpt},\cite{long2023discuss},\cite{chen2024mapgpt},\cite{lin2024correctable}.
For example, NavGPT~\cite{zhou2023navgpt} builds a purely GPT-4-based navigation agent that reasons the action decision based on the textual represented visual observations and navigational histories. DiscussNav~\cite{long2023discuss} incorporates multiple foundation models like InstructBLIP~\cite{dai2023instructblip} and GPT-4~\cite{OpenAI_2023} to address different navigation inputs and produce comprehensive reasoning for decision.

In this paper, we propose 
a rewriting-based VLN data augmentation approach driven by foundation models to mitigate the data scarcity issue of VLN and improve the generalization ability of navigation agents.
Rather than using foundation models for navigation decisions directly which need frequent queries, we adopt the foundation models for VLN data augmentation and therefore leading to a low-frequency query and reducing the cost significantly.  
Moreover, in contrast to existing zero-shot LLM-based VLN approaches
\cite{zhou2023navgpt},\cite{long2023discuss},\cite{chen2024mapgpt},
our approach effectively avoids the domain gap between foundation models and the VLN task through training supervised-learning-based navigation agents
\cite{Chen2021HistoryAM}, \cite{Chen2022ThinkGA} on both the human-annotated training data and our augmentation data.

\subsection{Data Augmentation in VLN}
Due to the collection difficulty of large-scale human-annotated trajectory-instruction pairs, data scarcity stands as a crucial challenge in VLN and researchers have adopted various augmentation strategies to tackle this issue. 
In this work, we mainly divide these approaches into two branches. The first branch is simulator-based
\cite{tan2019learning},\cite{fried2018speaker},\cite{liu2021vision},\cite{chen2022learning},\cite{wang2023scaling},\cite{Li2022EnvEditEE}, 
i.e., the augmentation data comes from the original Matterport3D~\cite{chang2017matterport3d} simulator or external simulators like HM3D~\cite{ramakrishnan2021habitat}. 
For example, Speaker-Follower~\cite{fried2018speaker} samples augmentation trajectories randomly in Matterport3D environment and trains a Speaker model to collect paired instructions. ScaleVLN~\cite{wang2023scaling} synthesizes 4.9 million trajectory-instruction pairs from  HM3D~\cite{ramakrishnan2021habitat} and Gibson~\cite{xia2018gibson} simulators.
However, the sampled trajectories are still 
constrained by specific simulator environments. 
Another branch is web-based, which collects large-scale images or videos from the web. AirBert~\cite{Guhur2021AirbertIP} introduces room images from Airbnb and Youtube-VLN~\cite{lin2023learning} collects room-tour videos on
YouTube for enhancing environmental diversity, respectively. Nevertheless, the noise of these web resources harms training and brings a heavy burden for data cleaning. 

In contrast to the above mentioned approaches, in this work, we propose to generate unseen observation-instruction pairs via rewriting human-annotated training data, fulfilling simulator-free and labor-saving
data 
augmentation for the VLN task.
A recent work~\cite{li2024panogen} also synthesizes new observations out of the simulator by introducing T2IMs, accompanied by Speaker-based instruction generation.
However, our work is different from it in two aspects: 
1) Our framework is rewriting-driven, where we rewrite observation descriptions and instructions to enable object-enriched observation generation and training-free instruction augmentation.
2) Our introduced mixing-then-focusing training strategy with the random observation cropping scheme effectively enhances the data distribution diversity while simultaneously restraining the negative impact of the augmentation data.

\begin{figure*}[t]
\begin{centering}
\includegraphics[width=\linewidth]{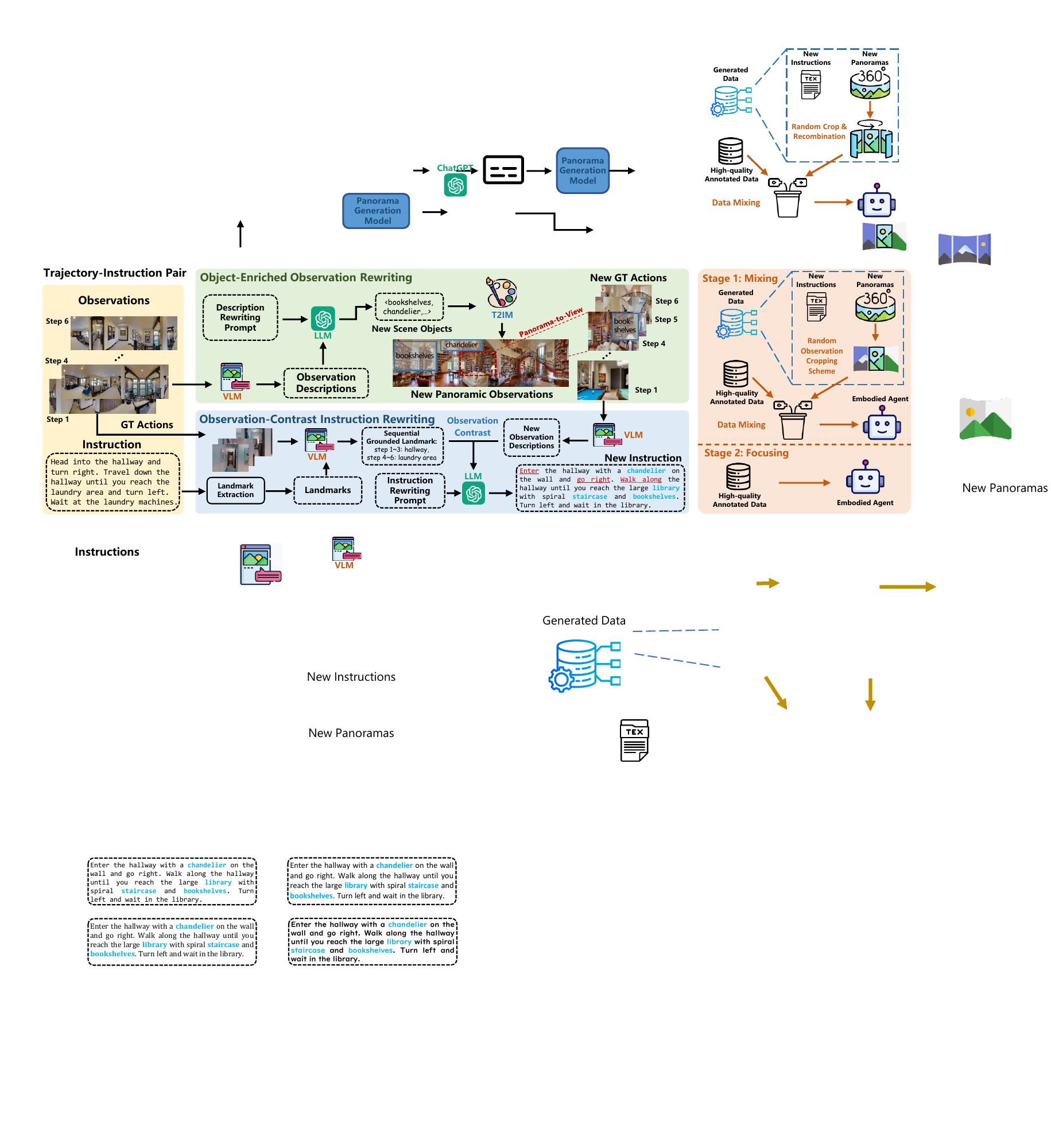}
\par\end{centering}
\caption{Overview of our \textbf{R}ewriting-driven \textbf{A}ug\textbf{M}entation (RAM) paradigm. 
For Object-Enriched Observation Rewriting, we collect object-enriched rewritten scene descriptions based on VLMs and LLMs. Then we feed the rewritten descriptions to T2IMs for synthesizing new observations via an efficient panorama-to-view scheme. 
During Observation-Contrast Instruction Rewriting, we ask the LLMs to perform observation contrast by reasoning the difference between original and new observation descriptions to generate new instructions. We further introduce a mixing-then-focusing strategy with a random observation cropping scheme for combining our rewritten trajectory-instruction pairs with human-annotated data for training. Newly generated objects and actional representations in the rewritten instruction are denoted in \textcolor{moti_blue}{\textbf{blue}} and \textcolor{moti_red}{\underline{red underlined}} fonts, respectively.}
\label{fig:overview}
\end{figure*}

\subsection{LLM-Driven Robotic Data Generation}
Thanks to the rich commonsense knowledge and powerful reasoning ability, LLMs have been proven a helpful data generation tool for alleviating the data shortage in robotic tasks by some recent works
\cite{xiao2022robotic},\cite{wang2023gensim},\cite{yang2024holodeck},\cite{wang2023robogen},\cite{zala2024envgen}.
DIAL~\cite{xiao2022robotic} employs GPT-3 to perform instruction augmentation by producing rephrased instructions guided by world knowledge. GenSim~\cite{wang2023gensim} uses GPT-4 to generate a task curriculum to enrich the existing benchmark by ten times. 
Holodeck~\cite{yang2024holodeck} leverages GPT-4 to generate spatial relational constraints between objects to create diverse scenes. RoboGen~\cite{wang2023robogen} utilizes GPT-4 to produce task proposals and scene configurations for realizing generative simulation. EnvGen~\cite{zala2024envgen} creates new training environments by prompting GPT-4 to generate a set of environment configurations which can be parsed by the simulator. 

In this work, we harness the physical world knowledge, reasoning ability, and language skills of the LLMs, and integrate the LLMs with other foundation models 
like T2IMs
for conducting observation-instruction rewriting for VLN.
As a result, our rewritten observations will have reasonable and diverse objects conforming to real-world scenarios. 
Moreover, the rewritten instructions will be more flexible and informative compared to Speaker-generated instructions and template-based instructions in previous VLN approaches.

\section{Method}
\label{method}
\newcommand{\hytt}[1]{\texttt{\hyphenchar\font=\defaulthyphenchar #1}}

In this section, we first describe the VLN problem setup (Section \textcolor{blue}{\ref{VLN Problem Setup}}) and then delve into the details of our RAM. 
The overview of RAM is illustrated in Fig.~\ref{fig:overview}. 
Specifically, given a human-annotated trajectory-instruction pair, RAM first performs Object-Enriched Observation Rewriting (Section \textcolor{blue}{\ref{Environment Rewriting}}), where Vision-Language Models (VLMs) and Large Language Models (LLMs) are introduced to generate object-enriched rewritten observation descriptions, based on which new observations are synthesized through an efficient panorama-to-view strategy via Text-to-Image Generation Models (T2IMs). Then, RAM conducts Observation-Contrast Instruction Rewriting (Section \textcolor{blue}{\ref{Instruction Rewriting}}), which collects grounded landmark information and observation descriptions respectively for the original and new trajectories, and asks the LLMs to contrast the difference between them to produce rewritten instructions. After collecting rewritten observation-instruction pairs, a two-stage mixing-then-focusing training strategy with a random observation cropping scheme is developed to effectively combine the rewritten data with the original data for training (Section \textcolor{blue}{\ref{mtftm}}). A notation summarization of variables appeared in this section is provided in Nomenclature.

\subsection{VLN Problem Setup}
\label{VLN Problem Setup}

In the discrete VLN task, 
an agent needs to explore the navigation connection graph $\mathcal{G}=(V,E)$, moving from a starting node to a target node following a given language instruction. $V$ and $E$ represent the nodes and edges in the navigation connectivity graph.
At timestep $t$, the agent receives a panoramic observation $O_{t}$ containing $K$ single-view observations $O_{t,k}$, i.e., $O_{t}=\{O_{t,k}\}_{k=1}^{K}$.
There are $N$ navigable views among $K$ views. The action space consists of the 
navigable views and a `STOP' action, and the agent chooses one as the action prediction $a_{t}$ from the action space at different timestep $t$. 


The continuous VLN is established on the Habitat~\cite{ramakrishnan2021habitat} simulator, where the agent can navigate to any points rather than the predefined graph nodes in the environment. Following existing works
\cite{wang2023scaling},\cite{Chen2021HistoryAM}, we adopt a pretrained waypoint predictor~\cite{hong2022bridging} to generate navigable waypoints in continuous environments, similar to that in the discrete VLN setting.

During VLN training, the agent is fed with trajectory-instruction pairs to learn cross-modal alignment knowledge and navigational skills, optimized by imitation learning algorithms or reinforcement learning algorithms
\cite{hong2021vln},\cite{Chen2021HistoryAM},\cite{Chen2022ThinkGA}.

\subsection{Object-Enriched Observation Rewriting}
\label{Environment Rewriting}

Many previous works
\cite{liu2021vision},\cite{Li2022EnvEditEE},\cite{li2024panogen}
focus on how to generate or introduce more environments for VLN training. For example,  EnvEdit~\cite{Li2022EnvEditEE} and REM~\cite{liu2021vision} edit the existing environments by changing room style and object appearances or mixing up environments. However, these approaches do not change spatial layouts among different objects
or introduce objects with new categories essentially. In our RAM, we propose Object-Enriched Observation Rewriting, which can extend the original scene to new scenes with diverse objects and spatial layouts efficiently for improving generalization. 

Object-Enriched Observation Rewriting consists of two primary procedures, namely Object-Enriched Scene Description Rewriting and Panorama-to-View Observation Generation. 
During Object-Enriched Scene Description Rewriting, we utilize VLMs to collect scene description $C_{t}$ for original observation $O_{t}$, then we leverage LLMs to provide object co-occurrence knowledge and generate rewritten object-enriched scene descriptions $C_{t}^{r}$. In Panorama-to-View Observation Generation, we obtain rewritten observations $O_{t}^{r}$ by feeding $C_{t}^{r}$ to T2IMs to directly synthesize rewritten panoramas, followed by a simple panorama discretization algorithm to produce new single-view observations. 

\noindent\textbf{Object-Enriched Scene Description Rewriting.}
For a human-annotated trajectory-instruction pair, we first use the Vision-Language Model (VLM) to collect panoramic observation descriptions during different navigation timesteps in the trajectory. 
For panoramic observation $O_{t}$ at timestep $t$, we get the scene description $C_{t}$ of $O_{t}$ by:
\begin{equation}
\label{eq:vlm}
C_{t}=\mathrm{VLM}\left(O_{t}\right).
\end{equation}
Then we construct the scene description rewriting prompt $P^{c}$, which contains the task definition formed as ``{\it Generate rewritten descriptions by adding the possible objects that may exist in the scene for the given scene description}''. 

\begin{figure}[t]
\begin{centering}
\includegraphics[width=\linewidth]{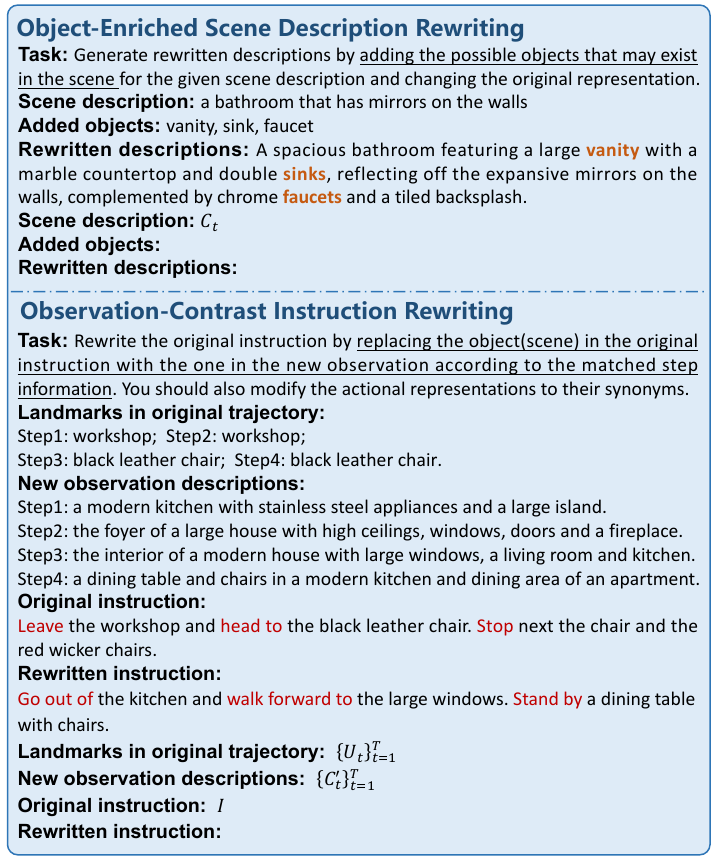}
\par\end{centering}
\caption{Prompts for Object-Enriched Scene Description Rewriting and Observation-Contrast Instruction Rewriting.}
\label{fig:prompt}
\end{figure}

To constrain the LLMs to explicitly indicate possible scene objects in its generated scene descriptions, we ask the LLMs in $P^{c}$ to generate both newly added objects $\{B_{t,n}\}_{n=1}^{N}$ and rewritten scene descriptions $C^{r}_{t}$ by feeding an in-context task example, as shown in Fig.~\ref{fig:prompt}. 
Additionally, to encourage the subsequent generated observations via T2IMs based on the rewritten scene descriptions $C^{r}_{t}$ to indicate different salient objects, which can enhance the diversity potentially, we also require the LLMs in $P^{c}$ to change the representation of original scene descriptions for highlighting different objects. The reason why we ask the LLMs to explicitly specify objects added to the synthetic scene is to facilitate adding objects with specific numbers as well as manually checking the generation quality of rewritten scene descriptions. And we do not utilize the outputs of added objects in the subsequent stages of our method.
Therefore, with our scene description rewriting prompt $P^{c}$ and the original scene description $C_{t}$, we obtain the rewritten object-enriched scene descriptions $C^{r}_{t}$ accompanied by added objects $\{B_{t,n}\}_{n=1}^{N}$ by:
\begin{equation}
C^{r}_{t}, \left\{B_{t,n}\right\}_{n=1}^{N}=\mathrm{LLM}\left(C_{t}, P^{c}\right),
\end{equation}
where $N$ is the number of added objects. 

\noindent\textbf{Panorama-to-View Observation Generation.}
We develop a panorama-to-view generation strategy for obtaining rewritten observations $O^{r}_{t}$, i.e., 
we first feed the rewritten object-enriched scene description $C^{r}_{t}$ to a panoramic T2IM once to collect rewritten panoramas, 
then we perform the panorama discretization to extract $K$ single-view observations, i.e., $O_{t}^{r}=\{O_{t,k}^{r}\}_{k=1}^{K}$. 
A recent work~\cite{li2024panogen} adopts a view-to-panorama generation manner, which needs to query the T2IM multiple times to generate each single-view observation, as well as stitch them into a panorama 
to improve view consistency. 
In contrast, our directly generated panoramas can naturally own view consistency.

For panorama generation, we choose MultiDiffusion~\cite{bar2023multidiffusion}, which is capable of generating images with desired aspect ratios and informative contents to generate the panoramic images.
We employ the Equirec2Perspec algorithm\footnote{  https://github.com/fuenwang/Equirec2Perspec} to discretize a panoramic image into 36 perspective images following the same setting in VLN~\cite{anderson2018vision}. 
Initially, the Equirec2Perspec algorithm utilizes the camera parameters, encompassing the Field of View (FOV), heading ($\theta$), and elevation ($\phi$), to calculate the inverse matrix of the camera’s projection matrix, denoted as $J_{inv}$, and the rotation matrix, denoted as $R$:
\begin{equation}
\begin{split}
J_{inv} &= \mathrm{MC_{proj}}(\mathrm{FOV}, \theta, \phi), \\ 
R &= \mathrm{MC_{rot}}(\theta, \phi),\\
\end{split}
\end{equation}
where $\mathrm{MC_{proj}}(\cdot)$ is the camera's projection matrix calculation function and $\mathrm{MC_{rot}}(\cdot)$ is the rotation matrix calculation function.

Then the algorithm takes $J_{inv}$, $R$, and multiple processing functions to convert pixel coordinates on the panoramic image $O^{r}_{t}$ to 3D point coordinates on the spherical surface, and then transform it to two-dimensional coordinates on the image plane to generate the single-view images $\{O^{r}_{t,k}\}_{k=1}^{K}$:
\begin{equation}
\begin{split}
\{O^{r}_{t,k}\}_{k=1}^{K} &= \mathrm{Equirec2Perspec}\left(J_{inv}, R, O^{r}_{t}\right),
\end{split}
\end{equation}
where the $\mathrm{Equirec2Perspec}$ function is the synthesis of a series of coordinate conversion functions. 
As shown in Fig.~\ref{fig:overview}, the generated single-view images (see ``New GT Actions'') have a similar field of view effect to the original single-view images in the Matterport3D~\cite{chang2017matterport3d} simulator.


\subsection{Observation-Contrast Instruction Rewriting}
\label{Instruction Rewriting}

As our newly generated panoramic observations of the trajectory are different from the original observations, the original paired instruction will not be aligned well with our new panoramic observations. In contrast to previous Speaker-based
\cite{fried2018speaker},\cite{Li2022EnvEditEE},\cite{li2024panogen}
or template-based
\cite{Guhur2021AirbertIP},\cite{lin2023learning}
instruction augmentation approaches, in this work, we introduce Observation-Contrast Instruction Rewriting to generate rewritten instructions $I^{r}$ based on the rewritten observations $\{O^{r}_{t}\}_{t=1}^{T}$, where $T$ is the
navigation step number for a specific trajectory. 

Specifically, we first perform Sequential Landmark Grounding, i.e., obtaining the grounded landmark $U_{t}$ for each observation $O_{t}$ in the original trajectory from the original instruction $I$. 
Then, we conduct New Observation Description Collection to derive descriptions $C'_{t}$ for the rewritten observations $O^{r}_{t}$.
Finally, we obtain rewritten instructions $I^{r}$ by asking the LLMs to: 
1) contrast each $U_{t}$ and $C'_{t}$ to replace $U_{t}$ in $I$ with the landmark appearing in $C'_{t}$ at the appropriate timestep, 
and 2) change the original actional representations in $I$ to their synonyms. 
Consequently,
our rewritten instructions can be observation-aligned with flexible representations.
Moreover, benefiting from rewriting human-annotated instructions, 
the generated instructions can be essentially more informative.

\noindent\textbf{Sequential Landmark Grounding.} 
Since the landmarks mentioned the instruction exist in the ground-truth action (observation) generally,  
for each ground-truth observation, we find its matched landmark in the instruction to perform sequential landmark grounding. Concretely, 
we first use an LLM to extract the sequential landmarks $U=\{U_{k}\}_{k=1}^{M}$ from the original instruction $I$ like that in~\cite{shahlm}, where $M$ is the size of the landmark list. 
Denote the ground-truth action (observation) at timestep $t$ extracted from the original observation $O_{t}$ as $G_{t}$. 
For each $G_{t}$, we employ a VLM~\cite{radford2021learning} to find its matched landmark $U_{t}$ by taking the landmark $U_{k}$ which has maximum similarity with $G_{t}$:
\begin{equation}
{U}_{t} = \mathop{\mathrm{argmax}}\limits_{U_{k}}\mathrm{Sim}\left(F^{v}\left(G_{t}\right), F^{t}\left(U_{k}\right) \right),
\vspace{-0.2cm}
\end{equation}
where $F^{t}$ and $F^{v}$ are the text encoder and the image encoder of the VLM, respectively. 
we represent the sequential grounded landmark information as $\{U_{t}\}_{t=1}^{T}$, where $T$ is the navigation step number for a specific trajectory.

\noindent\textbf{New Observation Description Collection.}
After sequential landmark grounding for each ground-truth action (observation) $G_{t}$ in the original trajectory to collect $\{U_{t}\}_{t=1}^{T}$, 
we extract the ground-truth action (observation) $G'_{t}$ which has the same position as $G_{t}$ from the rewritten observations $O_{t}^{r}$. Then, we adopt the VLM used during Object-Enriched Scene Description Writing to generate the 
description $C'_{t}$ for the ground-truth action (observation) $G'_{t}$ like Eq.\ref{eq:vlm}.
We represent the sequential observation descriptions for new ground-truth observation sequence $\{G'_{t}\}_{t=1}^{T}$ 
as $\{C'_{t}\}_{t=1}^{T}$.

\noindent\textbf{Instruction Rewriting by Observation Contrast.}
As shown in Fig.~\ref{fig:prompt}, we build the instruction rewriting prompt $P^{i}$ with the task description of ``{\it Rewrite the original instruction by replacing the object (scene) in the original instruction with the one in the new observation...}'',  followed by an in-context task example, which teaches the LLMs to perform observation contrast as well as change the representations of actional descriptions to produce rewritten instructions.
Therefore, with our instruction rewriting prompt $P^{i}$, the grounded landmark information $\{U_{t}\}_{t=1}^{T}$, the new observation descriptions $\{C'_{t}\}_{t=1}^{T}$, and the original instruction $I$, 
the rewritten instruction $I^{r}$ is acquired by:
\begin{equation}
I^{r} = \mathrm{LLM}(\{U_{t}\}_{t=1}^{T}, \{C'_{t}\}_{t=1}^{T},  I, P^{i})
\end{equation}

\subsection{Mixing-then-Focusing Training Mechanism}
\label{mtftm}

After obtaining the rewritten data, one simple way is to mix it with the original data for training directly. However, due to the inherent deficiency of foundation models, e.g., the repeated object generation issue of the panoramic T2IMs~\cite{Cai2024LMAGICLM}, arbitrarily introducing rewritten data may bring unexpected noise. 
Therefore, we introduce an effective two-stage mixing-then-focusing training mechanism, 
where we mix the original data with rewritten data in Stage 1, and use pure original data to reduce the noise impact in Stage 2. During  stage 1, we additionally introduce a random observation cropping scheme to serve as a data augmentation method, i.e., for a rewritten panorama, we randomly crop it into large patches, and combine the patches from different panoramas to form new panoramas. This further strengthens the diversity of rewritten data while mitigating the repeated object generation issue of the T2IMs. Denote the original training data as $\{\{O_{t}\}_{t=1}^{T},I\}$ and our rewritten data as $\{\{O^{r}_{t}\}_{t=1}^{T},I^{r}\}$, where $T$ is the step number of a trajectory. The navigation losses  for the two stages $\mathcal{L}_{\mathrm{s1}}$ and $\mathcal{L}_{\mathrm{s2}}$ are calculated respectively by:
\begin{equation}
\mathcal{L}_{\mathrm{s1}} = E^{n}\left(\left\{\left\{O_{t}\right\}_{t=1}^{T},I\right\},\left\{\mathrm{RC}(\left\{O^{r}_{t}\right\}_{t=1}^{T}),I^{r}\right\}\right),
\end{equation}
\begin{equation}
\mathcal{L}_{\mathrm{s2}} = E^{n}\left(\left\{\left\{O_{t}\right\}_{t=1}^{T},I\right\}\right),
\end{equation}
where $E^{n}$ represents the navigation agent and $\mathrm{RC}(\cdot)$ denotes our random observation cropping scheme.

\section{Experiments}
\label{experiments}

\subsection{Experimental Setup}

\subsubsection{Datasets}

We mainly evaluate our RAM on three discrete VLN benchmarks: R2R~\cite{anderson2018vision},  REVERIE~\cite{qi2020reverie}, and  R4R~\cite{jain2019stay}. 
R2R contains 90 indoor scenes with 7189 trajectories. 
Each trajectory is accompanied by 3 step-by-step instructions. On average, an instruction contains 32 words, and each ground-truth path is
formed by seven nodes with a total length of 10$m$.
REVERIE replaces the fine-grained instructions in R2R with high-level instructions that are only related to the target position and object.
R4R creates longer instructions and trajectories by concatenating two adjacent tail-to-head trajectories in R2R. 
Both R2R and REVERIE split the evaluation set into Val Seen, Val Unseen, and Test Unseen, while R4R only splits the evaluation set into Val Seen and Val Unseen. ``Val Seen'' means new trajectory-instruction pair data but collected from the same houses as those in the training dataset. On the other hand, ``Val Unseen'' and ``Test Unseen'' stand for entirely different VLN data collected from additional houses which are different from the houses in the training dataset.

We also extend our RAM to R2R-CE~\cite{krantz2020beyond}, which transfers the discrete trajectories in R2R
to continuous 3D scans rendered by Habitat~\cite{savva2019habitat} simulator. The R2R-CE dataset
contains 16k instruction-trajectory pairs after removing non-transferable paths.

\subsubsection{Evaluation Metrics}
We use the following metrics for evaluation 
on R2R and R2R-CE:
1) Trajectory Length (TL): the average length of the agent's navigated path in meters, 2) Navigation Error (NE): the average distance in meters between the agent's destination and the target viewpoint, 3) Success Rate (SR): the ratio of success, where the agent stops within three meters of the target point, and 4) Success rate weighted by Path Length (SPL): success rate normalized by the ratio between the length of the shortest path and the predicted path. 
Three other metrics are used for REVERIE:
5) Remote Grounding Success Rate (RGS): the ratio of grounding the correct object when stopping, 6) Remote Grounding Success rate weighted by Path Length (RGSPL): weight RGS by TL, and 7) Oracle Success Rate (OSR): the ratio of containing a viewpoint along the path where the target object is visible. For R4R, we further adopt 8) the Coverage weighted by Length Score (CLS), 9) the normalized Dynamic Time Warping (nDTW), and 10) the Success weighted by nDTW (SDTW) for measuring the path fidelity.

\begin{table*}[t]
	\fontsize{12}{12}\selectfont

\caption{Comparison with existing methods on R2R. *: 900 extra simulator scenes for training and using CLIP ViT-H/14 as image features. \dag: using CLIP ViT-B/16 as image features. \ddag: using CLIP ViT-L/14 as image features.}
	\label{tab1:com_with_sota_r2r}
	\resizebox{1.0\linewidth}{!}{
	{\renewcommand{\arraystretch}{1.2}
		\begin{tabular}{c||c|c|c|c|c|c|c|c|c|c|c|c}

			\specialrule{.1em}{.05em}{.05em}
			\multirow{2}{*}{Method}&\multicolumn{4}{c|}{Val Seen }&\multicolumn{4}{c|}{Val Unseen}&\multicolumn{4}{c}{Test Unseen}\cr\cline{2-13}
			&TL&NE $\downarrow$&SR $\uparrow$&SPL $\uparrow$&TL&NE $\downarrow$&SR $\uparrow$&SPL $\uparrow$&TL&NE $\downarrow$&SR $\uparrow$&SPL $\uparrow$\cr
			\hline
			
        
            
            ScaleVLN*~\cite{wang2023scaling}&13.24&2.12&81&75&14.09&2.09&81&70&13.93&2.27&80&70\\
            \hline
            Seq2Seq~\cite{anderson2018vision}&11.33&6.01&39&-&8.39&7.81&22&-&8.13&7.85&20&18\\
            Speaker-Follower~\cite{fried2018speaker}&-&3.36&66&-&-&6.62&35&-&14.82&6.62&35&28\\
            
   
          EnvDropout~\cite{tan2019learning}&11.00&3.99&62&59&10.70&5.22&52&48&11.66&5.23&51&47\\  
   Airbert~\cite{Guhur2021AirbertIP}&11.09&2.68&75&70&11.78&4.01&62&56&12.41&4.13&62&57\\

   HAMT~\cite{Chen2021HistoryAM} &11.15&2.51&76&72&11.46&3.62&66&61&12.27&3.93&65&60\\ 
    EnvEdit~\cite{Li2022EnvEditEE}&11.18&2.32&77&74&11.13&3.24&69&64&11.90&3.59&68&64\\
    YouTube-VLN~\cite{lin2023learning}&-&-&-&-&14.58&2.90&74&62&16.13&3.44&72&60\\

    PanoGen~\cite{li2024panogen}&-&-&-&-&13.40&3.03&74&64&14.38&3.31&72&62\\
LSAL~\cite{10414007}&11.19&2.88&73&70&12.43&3.62&65&59&13.51&4.00&63&58\\

SS3DSRL~\cite{tan2024self}&-&2.35&80&73&-&3.36&71&58&-&3.73&70&58\\

AZHP~\cite{10531078}&12.38&2.08&80&74&13.68&3.25&71&60&14.47&3.43&70&59\\

LANA~\cite{10359152}&-&-&-&-&13.50&-&73&61&14.50&-&70&60\\

    \hline
    DUET$^{\dag}$~\cite{Chen2022ThinkGA} (baseline)&13.86&2.12&80.80&73.64&16.22&3.06&72.37&58.75&-&-&-&-\\
    RAM$^{\dag}$(ours)&11.50&1.95&82.17&77.70&13.33&2.96&73.65&63.13&14.69&3.34&71&61\\
    DUET$^{\ddag}$~\cite{Chen2022ThinkGA} (baseline)&11.64&1.92&\textbf{82.76}&\textbf{78.17}&13.05&3.07&73.22&64.49&-&-&-&-\\
    %
    RAM$^{\ddag}$(ours)&11.97&\textbf{1.85}&82.47&77.98&13.39&\textbf{2.66}&\textbf{76.29}&\textbf{66.39}&14.50&\textbf{3.08}&\textbf{75}&\textbf{65}\\
 \specialrule{.1em}{.05em}{.05em}

		\end{tabular}}}
\end{table*}

\begin{table*}[t]
	\fontsize{12}{12}\selectfont
 \centering
\caption{Navigation and object grounding performance on REVERIE. *: 900 extra simulator scenes for training. \ddag: using CLIP ViT L/14 as image features.}
	\label{tab:com with sota on reverie}
	\resizebox{\linewidth}{!}{
	{\renewcommand{\arraystretch}{1.2}
		\begin{tabular}{c||c|c|c|c|c|c|c|c|c|c|c|c}

			\specialrule{.1em}{.05em}{.05em}
    \multirow{3}{*}{Method}
    &\multicolumn{6}{c|}{\textbf{Val Unseen}}&\multicolumn{6}{c}{\textbf{Test Unseen}}\cr\cline{2-13}&\multicolumn{4}{c|}{Navigation}&\multicolumn{2}{c|}{Grounding}&\multicolumn{4}{c|}{Navigation}&\multicolumn{2}{c}{Grounding}\cr\cline{2-13}
			&TL&SR $\uparrow$&OSR$\uparrow$&SPL $\uparrow$&RGS$\uparrow$&RGSPL$\uparrow$&TL&SR $\uparrow$&OSR$\uparrow$&SPL $\uparrow$&RGS$\uparrow$&RGSPL$\uparrow$\cr
			\hline
			
        
            AutoVLN*~\cite{chen2022learning}&-&55.89&62.14&40.85&36.58&26.76&-&55.17&62.30&38.88&32.23&22.68\\
            ScaleVLN*~\cite{wang2023scaling}&-&56.97&63.85&41.84&35.76&26.05&-&56.13&62.65&39.52&32.53&22.78\\
            \hline
            RCM~\cite{wang2019reinforced}&11.98&9.29&14.23&6.97&4.89&3.89&10.60&7.84&11.68&6.67&3.67&3.14\\
           
          SMNA~\cite{ma2019self}&9.07&8.15&11.28&6.44&4.54&3.61&9.23&5.80&8.39&4.53&3.10&2.39\\
         FAST-MATTN~\cite{qi2020reverie}&45.28&14.40&28.20&7.19&7.84&4.67&39.05&19.88&30.63&11.60&11.28&6.08\\  
    SIA~\cite{Lin2021SceneIntuitiveAF}&41.53&31.53&44.67&16.28&22.41&11.56&48.61&30.80&44.56&14.85&19.02&9.20\\
    VLN$\circlearrowright$BERT~\cite{hong2021vln}&16.78&30.67&35.02&24.90&18.77&15.27&15.86&29.61&32.91&23.99&16.50&13.51\\
    
    Airbert~\cite{Guhur2021AirbertIP}&18.71&27.89&34.51&21.88&18.23&14.18&17.91&30.28&34.20&23.61&16.83&13.28\\    
    DUET~\cite{Chen2022ThinkGA} &22.11&46.98&51.07&33.73&32.15&23.03&21.30&52.51&56.91&36.06&31.88&22.06\\
    KERM~\cite{Li2023KERMKE}&21.85&49.02&53.65&34.83&33.97&\textbf{24.14}&17.32&52.26&57.44&37.46&32.69&23.15\\
    YouTube-VLN~\cite{lin2023learning}&21.87&48.11&53.71&34.43&32.15&23.43&21.94&54.32&60.51&37.34&32.02&21.94\\

AZHP~\cite{10531078}&22.08&49.02&53.88&\textbf{36.25}&32.41&24.13&21.10&52.52&57.39&36.11&32.10&22.54\\
    
    \hline
    DUET$^{\ddag}$~\cite{Chen2022ThinkGA} (baseline)&25.64&48.71&53.62&34.26&32.18&22.64&-&-&-&-&-&-\\
RAM$^{\ddag}$(ours)&25.44&\textbf{51.89}&\textbf{58.47}&35.00&\textbf{34.31}&23.20&22.78&\textbf{57.44}&\textbf{64.26}&\textbf{41.41}&\textbf{36.05}&\textbf{25.77}\\

 \specialrule{.1em}{.05em}{.05em}

			\end{tabular}}}
\end{table*}

\subsubsection{Implementation Details}

We plug our RAM into a powerful pretraining-based baseline~\cite{Chen2022ThinkGA}. In the pretraining phase, the agent is trained with three proxy tasks: Masked Language Modeling (MLM), Instruction and Trajectory Matching (ITM), and Single Action Prediction
(SAP). 
In the finetuning phase, the agent is trained with DAGGER~\cite{ross2011reduction} method. 
We add our rewritten data during both the pretraining and finetuning phases to boost the navigation training. 
The pretraining is conducted for 20k iterations with a batch size of 128 and learning rate of 5e-5 on four NVIDIA RTX 3090 GPUs. 
Our mixing-then-focusing training strategy is implemented during the finetuning phase, where the maximum training iterations, the batch size, and the learning rate for two stages are set as 20k, 8, and 1e-5, respectively. We resume the best model in stage 1 to train for stage 2. 
The finetuning is implemented on a single NVIDIA RTX 3090 GPU. 
The pretraining phase lasts about 50 hours and the finetuning phase lasts about 20 hours.

We adopt two kinds of visual encoders, i.e., CLIP ViT-B/16 and CLIP ViT-L/14~\cite{radford2021learning} to validate the effectiveness of our method. 
Like previous works
\cite{Chen2021HistoryAM},\cite{Chen2022ThinkGA},\cite{li2024panogen}, we also add the prevalent data~\cite{hao2020towards} during training. 
We use Tag2Text~\cite{huang2023tag2text} as the VLM for observation description collection, and employ ChatGPT~\cite{openai}(with version \textit{gpt-3.5-turbo-1106}) as the LLM for scene description rewriting and instruction rewriting. 
For an original trajectory-instruction pair, we collect 3 different trajectories containing new panoramas, and for each new trajectory we obtain one corresponding augmentation instruction.
The LLMs, VLMs, T2IMs used in the experiments are all in a zero-shot manner without the need for training consumption. 
The parameter size of our model is 688.51M.

\noindent\textbf{Transfer to Continuous Setting.}
Following previous works like ScaleVLN~\cite{wang2023scaling}, we transfer our RAM to the continuous VLN setting by first pretraining the baseline model DUET~\cite{Chen2022ThinkGA} with both original and our rewritten observation-instruction data on the R2R discrete environments. Then we finetune the pretrained model on R2R-CE based on a pretrained candidate waypoint predictor~\cite{hong2022bridging} for a fair comparison with other methods. The finetuning phase is conducted for 30 epochs with a batch size of 8 and a learning rate of 1e-5 on two NVIDIA RTX 3090 GPUs.

\definecolor{vis_red}{RGB}{255,0,0}
\definecolor{vis_blue}{RGB}{31,78,121}

\subsection{Comparison with Existing Approaches}

\noindent\textbf{R2R.} 
Table~\ref{tab1:com_with_sota_r2r} shows the performance comparison of different methods on R2R, where we can find that our RAM using CLIP ViT-L/14 features outperforms all previous works that do not introduce large-scale additional data for training like ScaleVLN~\cite{wang2023scaling}. Although  ScaleVLN outperforms previous works as well as RAM, it resorts to significantly larger realistic data resources ($\times$352 larger than the R2R dataset) while our RAM only introduces a small scale of generated data ($\times$3 larger than the R2R dataset).
Moreover, our RAM with both two kinds of CLIP features outperforms ScaleVLN~\cite{wang2023scaling} in Val Seen split. 
We can also observe that our RAM outperforms baseline DUET~\cite{Chen2022ThinkGA} especially in Val Unseen when using both CLIP ViT-B/16 and ViT-L/14 as image features, revealing its superior generalization ability.

\noindent\textbf{REVERIE.} 
Table~\ref{tab:com with sota on reverie} compares different methods regarding the navigation and object grounding performance on REVERIE, where we can find that RAM surpasses the baseline~\cite{Chen2022ThinkGA} largely on both navigation and object grounding metrics under the same visual features on Val Unseen, e.g., the improvements for SR and RGS are $\sim$3.1\% and $\sim$2.2\%, respectively.
RAM also significantly outperforms previous works including AutoVLN~\cite{chen2022learning} and ScaleVLN~\cite{wang2023scaling} that introduce large-scale additional simulator data on Test Unseen regarding both navigation and object grounding metrics.
These results show that our RAM effectively improves the generalization without the need to introduce large-scale additional realistic data. 

\begin{table}[t]
	\fontsize{20}{20}\selectfont
	 \caption{Performance comparison on R4R dataset. \ddag: using CLIP ViT-L/14 as image features.}	
  \resizebox{1.0\linewidth}{!}{
	{\renewcommand{\arraystretch}{1.2}
		\begin{tabular}{c||c|c|c|c|c|c}

			\specialrule{.1em}{.05em}{.05em}
			\multirow{2}{*}{Model}&\multicolumn{6}{c}{Val Unseen}\cr\cline{2-7}
			&NE$\downarrow$&SR$\uparrow$&SPL$\uparrow$&nDTW$\uparrow$&sDTW $\uparrow$&CLS $\uparrow$\cr
			\hline
			
        

            SF~\cite{fried2018speaker}&8.47&24&-&-&-&30\\
   RCM~\cite{wang2019reinforced}&-&29&-&30&13&35\\
   PTA~\cite{landi2019perceive}&8.25&24&-&32&10&37\\
   EGP~\cite{deng2020evolving}&8.0&30.2&-&37.4&17.5&44.4\\

   
            DUET-CLIP~\cite{li2024panogen}&6.06&46.61&41.94&-&-&-\\
            PanoGen~\cite{li2024panogen}&6.02&47.78&44.25&-&-&-\\
            \hline
            DUET\ddag~\cite{Chen2022ThinkGA} (baseline)&5.88&50.35&46.14&40.08&27.91&46.31\\
            RAM\ddag (ours)&\textbf{5.18}&\textbf{55.28}&\textbf{49.59}&\textbf{42.05}&\textbf{29.91}&\textbf{47.18}\\
 \specialrule{.1em}{.05em}{.05em}

		\end{tabular}}}

  	\label{tab:r4r}

\end{table}

\begin{table}
	\fontsize{18}{18}\selectfont
	\caption{Results on R2R-CE dataset. $\dag$:
using the same waypoint predictor~\cite{hong2022bridging} for a fair comparison.}	
	\label{tab:r2r-ce}
	\centering
			\resizebox{\linewidth}{!}{
	{\renewcommand{\arraystretch}{1.2}\begin{tabular}{c||c|c|c|c|c|c}
			 \specialrule{.1em}{.05em}{.05em}
		 \multirow{2}{*}{Method}&\multicolumn{3}{c|}{Val Seen}&\multicolumn{3}{c}{Val Unseen}\cr\cline{2-7}
    &NE$\downarrow$&OSR$\uparrow$&SR$\uparrow$&NE$\downarrow$&OSR$\uparrow$&SR$\uparrow$\cr\hline
        CMTP~\cite{chen2021topological}&7.10&45.4&36.1&7.9&38.0&26.4\\
        LAW~\cite{raychaudhuri2021language}&6.35&49&40&6.83&44&35\\
        Sim2Sim~\cite{krantz2022sim}&4.67&61&\textbf{52}&6.07&52&43\\
        CMA$^{\dag}$~\cite{hong2022bridging}&5.20&61&51&6.20&52&41\\
        VLN-RecBERT$^{\dag}$~\cite{hong2022bridging}&5.02&59&50&5.74&53&44\\
        \hline
          DUET$^{\dag}$~\cite{Chen2022ThinkGA} (baseline)&4.46&62.98&43.32&5.14&59.65&37.25\\
          RAM$^{\dag}$(ours)&\textbf{4.09}&\textbf{65.68}&46.92&\textbf{4.95}&\textbf{61.45}&\textbf{44.15}\\

        
          \specialrule{.1em}{.05em}{.05em}
		\end{tabular}}}
\vspace{-0.4cm}
\end{table}

\noindent\textbf{R4R.}
Table~\ref{tab:r4r} presents the results on R4R, where RAM outperforms the baseline~\cite{Chen2022ThinkGA} in both navigation (NE, SR, and SPL) and instruction following (nDTW, sDTW,  and CLS) metrics on Val Unseen, showing the effectiveness of our method when facing much longer instructions and trajectories. 
It is also superior to a recent data augmentation method~\cite{li2024panogen} that utilizes T2IMs for environment generation based on the original scene description and adopts Speaker-based instruction augmentation. 
This comparison result shows that our object-enriched observation rewriting strategy can better improve the scene diversity by introducing new objects and spatial layouts. The comparison also demonstrates the superiority of our observation-contrast instruction rewriting scheme over Speaker-based instruction augmentation.  


\noindent\textbf{Transfer to R2R-CE.}
We further transfer our RAM to R2R-CE. As shown in Table \ref{tab:r2r-ce}, although our work is mainly established based on discrete environments, it still shows impressive superiority over the baseline method~\cite{Chen2022ThinkGA} on R2R-CE, revealing its potential to generalize to continuous environments which is more realistic. 


\begin{table}[t]
	\fontsize{12}{12}\selectfont
	 \caption{Ablation results for observation-instruction rewriting. ``Pan'' stands for generated panoramic images, ``Des.'' means rewritten object-enriched scene descriptions,  and ``Ins.'' means  rewritten instructions, respectively.}
 \resizebox{1.0\linewidth}{!}{
	{\renewcommand{\arraystretch}{1.2}
		\begin{tabular}{c||c|c|c|c|c|c|c}

			\specialrule{.1em}{.05em}{.05em}
			\multirow{2}{*}{No.}&\multicolumn{3}{c|}{Settings}&\multicolumn{2}{c|}{Val Seen}&\multicolumn{2}{c}{Val Unseen}\cr\cline{2-8}
			&Pan.&Des.&Ins.&NE $\downarrow$&SR $\uparrow$&NE $\downarrow$&SR $\uparrow$\cr
			\hline
			
        
            baseline&\ding{55}&\ding{55}&\ding{55}&2.91&74.73&3.70&65.94\\
            speaker inst&\ding{55}&\ding{55}&\ding{55}&3.19&70.91&3.83&65.05\\
            \hline
            1&\ding{51}&\ding{55}&\ding{55}&2.76&74.14&3.54&67.52\\
            2&\ding{55}&\ding{55}&\ding{51}&2.76&74.83&3.66&67.43\\
            
            3&\ding{51}&\ding{55}&\ding{51}&2.66&74.34&3.51&68.08\\
            4&\ding{51}&\ding{51}&\ding{55}&2.44&77.08&3.50&69.69\\
            5&\ding{51}&\ding{51}&\ding{51}&\textbf{2.37}&\textbf{78.55}&\textbf{3.42}&\textbf{70.29}\\
 \specialrule{.1em}{.05em}{.05em}

		\end{tabular}}}
  \vspace{-0.4cm}
 
  	\label{tab:component ablation}

\end{table}

\begin{table}[t]
	\fontsize{12}{12}\selectfont

	 \caption{Ablation results of adding data augmented via RAM in different training phases on R2R. }
 \resizebox{1.0\linewidth}{!}{
	{\renewcommand{\arraystretch}{1.4}
		\begin{tabular}{c||c|c|c|c|c|c}

			\specialrule{.1em}{.05em}{.05em}
			\multirow{2}{*}{No.}&\multicolumn{2}{c|}{Settings}&\multicolumn{2}{c|}{Val Seen}&\multicolumn{2}{c}{Val Unseen}\cr\cline{2-7}
			&Pretraining&Finetuning&SR $\uparrow$&SPL $\uparrow$&SR $\uparrow$&SPL $\uparrow$\cr
			\hline
			
        
            baseline&\ding{55}&\ding{55}&80.80&73.64&72.37&58.75\\
            
            \hline
            1&\ding{51}&\ding{55}&80.71&75.08&71.99&62.32\\
            2&\ding{55}&\ding{51}&80.71&74.56&73.61&62.46\\
            
3&\ding{51}&\ding{51}&\textbf{82.17}&\textbf{77.70}&\textbf{73.65}&\textbf{63.13}\\
 \specialrule{.1em}{.05em}{.05em}

		\end{tabular}}}
 
  	\label{tab:training ablation}

	\vspace{-0.2cm}
\end{table}

\begin{figure}[!htb]
\begin{centering}
\includegraphics[width=\linewidth]
{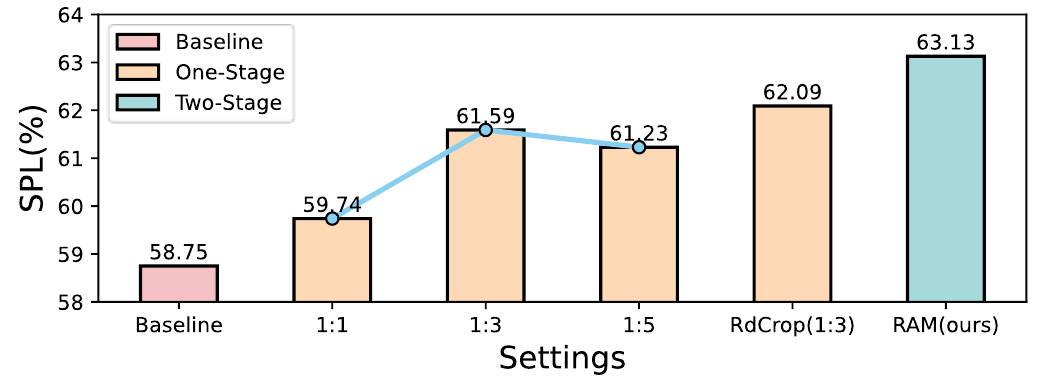}
\par\end{centering}
\caption{Ablation results for mixed training strategy on R2R dataset. 
``1:1'', ``1:3'', and ``1:5'' represent the data mixing ratio between the original human-annotated data and our rewritten data. ``RdCrop(1:3)'' means using our random observation cropping scheme with the data mixing ratio of 1:3. 
}
\label{fig:mix training}
\end{figure}

\subsection{Ablation Study}
We conduct ablation studies to analyze the effects of different method components and the impacts of adding our augmentation data during different training phases. We use CLIP ViT-B/16 features for the ablation experiments.
In Table~\ref{tab:component ablation}, since we aim to verify the effectiveness of our rewritten data and exclude the effect of high-quality human annotated instructions, we train the baseline using the widely used R2R augmentation dataset PREVALENT~\cite{hao2020towards}, which can  provide a fair comparison of different kinds of augmentation data.
For the baseline in Table~\ref{tab1:com_with_sota_r2r},  Fig.~\ref{fig:mix training}, and Table~\ref{tab:training ablation}, we reproduce the baseline model DUET~\cite{Chen2022ThinkGA} with CLIP VIT-B/16 visual features.

\begin{figure*}[t]
\begin{centering}
\includegraphics[width=\linewidth]{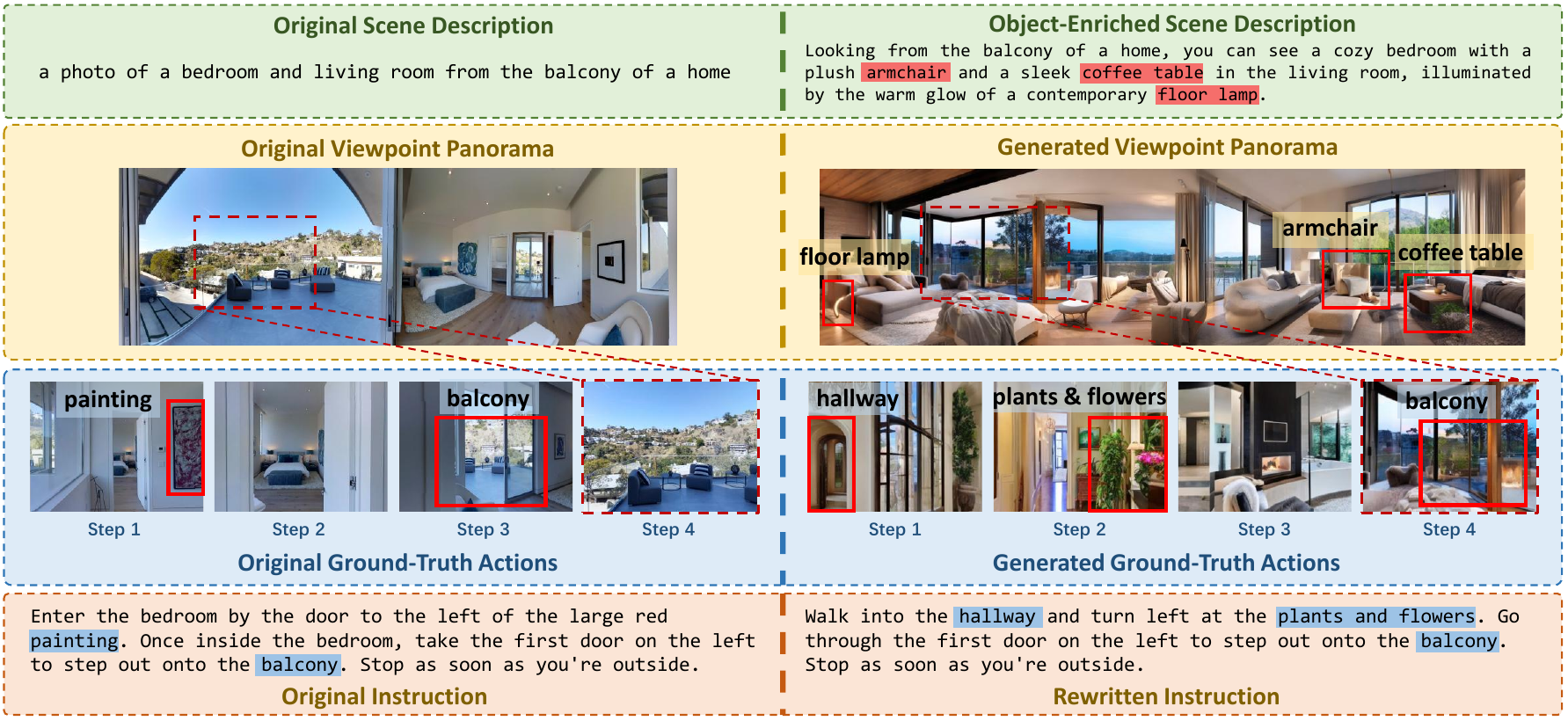}
\par\end{centering}
\caption{
Visualization examples of rewritten object-enriched scene description, generated panorama, extracted ground-truth sequence from the generated panorama, and the rewritten instruction. Newly generated scene objects and modality-aligned objects in the instruction are denoted in \textcolor{vis_red}{red} boxes and \textbf{bold} fonts, respectively.
}\label{fig_visual}
\vspace{-0.4cm}
\end{figure*}

\begin{table}[t]
    \centering
    \fontsize{12}{12}\selectfont
    \caption{Comparison with the baseline method~\cite{Chen2022ThinkGA} under the low-resource settings. Using CLIP ViT-B/16 as image features.}
    \resizebox{1.0\linewidth}{!}{
    {\renewcommand{\arraystretch}{1.2}
    \begin{tabular}{c||cc|cc|cc}
    \specialrule{.1em}{.05em}{.05em}
        \multirow{3}{*}{Model}&\multicolumn{6}{c}{Val Unseen}\\\cline{2-7}
    
        & \multicolumn{2}{c|}{20\%}
        & \multicolumn{2}{c|}{40\%}
        & \multicolumn{2}{c}{60\%}\\
    \cline{2-7}
          & NE$\downarrow$ & SR$\uparrow$ & NE$\downarrow$ & SR$\uparrow$ & NE$\downarrow$ & SR$\uparrow$   \\
    \hline
        baseline~\cite{Chen2022ThinkGA}&3.58&68.03&3.26&71.05&3.43&69.39\\
        RAM (ours)&\textbf{3.36}&\textbf{70.80}&\textbf{3.19}&\textbf{71.99}&\textbf{3.19}&\textbf{72.71}\\

    \specialrule{.1em}{.05em}{.05em}
    \end{tabular}}}
   \vspace{-0.4cm} 
   \label{tab:low resource}
\end{table}

\noindent\textbf{Effectiveness of Observation-Instruction Rewriting.}
Table~\ref{tab:component ablation}
presents the ablation results of our observation-instruction rewriting, where we can draw the following crucial conclusions: (1) By comparing No.1$\sim$No.5 with the baseline and speaker inst, we can find that both our rewritten observations and instructions augmented by RAM are helpful for increasing the diversity of training data, leading to large performance improvement. For example, No.1 and No.2 which perform solely observation or instruction rewriting have already outperformed the baseline. Moreover, No.5 with both rewriting procedures achieves the best performance. (2) The comparison between No.1 and No.4 shows the effectiveness of object-enriched description rewriting that generating observations based on the rewritten scene descriptions brings larger performance enhancement than that based on the original scene description. 
(3) The superior performance of No.5 compared to No.3 (Pan. \& Ins.) indicates that our object-enriched observation rewriting scheme effectively enhances observation diversity through the object-enriched scene description rewriting strategy (denoted as ``Des.''), thereby leading to significant performance improvements.
(4) No.2 significantly outperforms speaker inst that using Speaker-augmented instructions,  revealing that our rewritten instructions are more informative and diverse, which are beneficial for improving the cross-modal alignment ability of the VLN agent.

\noindent\textbf{Effectiveness of Mixing-then-Focusing Training Mechanism.}
Fig.~\ref{fig:mix training} presents the ablation results for different data fusion strategies for finetuning training. The results of RAM {\it vs.} one-stage variant (denoted as 1:1, 1:3, 1:5) 
sufficiently show the effectiveness of our mixing-then-focusing training strategy. 
The improvement of RdCrop (1:3) over ``1:3'' further shows the effectiveness of our random observation cropping scheme that mitigates the noise brought by the foundation models and improves the data distribution diversity. 
Moreover, by comparing mixing training with different proportions of rewritten data (1:1, 1:3, 1:5), we can find that simply adding more rewritten data may not contribute to linear performance improvement. This reveals the importance of exploring effective data fusion mechanisms for activating the advantage of augmentation data, and our mixing-then-focusing training strategy can inspire future research for this. 

\noindent\textbf{Pretraining \& Finetuning Ablation Results.}
Table~\ref{tab:training ablation} gives the ablation results of adding RAM augmentation data in different training phases. 
For No.1 in Table~\ref{tab:training ablation}, we directly mix the original data with our rewritten data in the pretraining phase, and then finetune the model using only the original data. For No.2 in Table~\ref{tab:training ablation}, we conduct the pretraining following the baseline model using original data, and then implement our mixing-then-focusing training mechanism using both original and rewritten data in the finetuning phase.
In Table \ref{tab:training ablation}, through the comparison between the baseline and No.2,  we can find that adding our RAM data during the finetuning phase can effectively improve the navigation performance in Val Unseen, showing that our augmentation data is helpful for improving the generalization ability of the VLN agent to unseen scenarios. Moreover, No.3 shows that by further adding our augmentation data in the pretraining phase, the metric results under both Val Seen and Val Unseen can be further boosted. These results show the positive impact of our RAM for enhancing training in different training phases.


\begin{figure*}[!htb]
\begin{centering}
\includegraphics[width=0.95\linewidth]{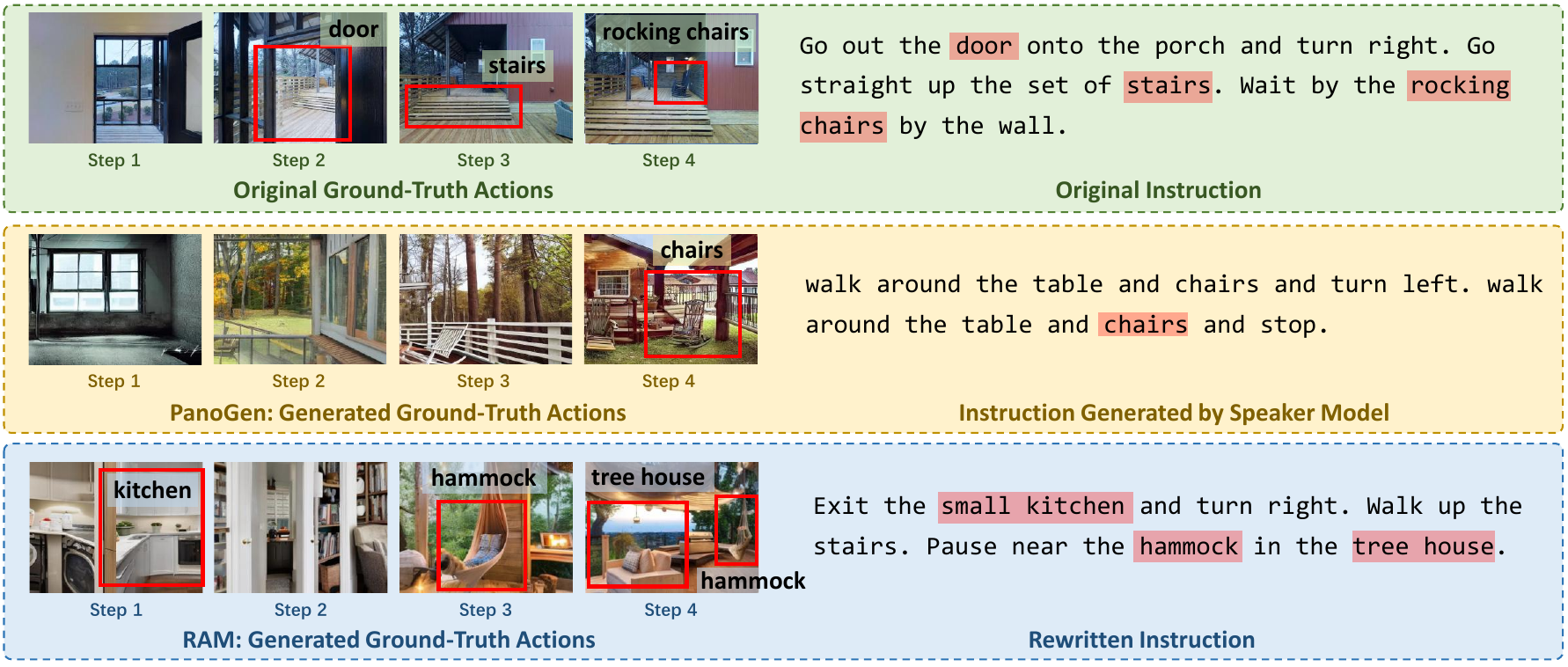}
\par\end{centering}
\vspace{-0.2cm}
\caption{Visualization comparison of instructions generated by the Speaker model~\cite{li2024panogen} and our RAM. Modality-aligned objects in the instructions are denoted in \textcolor{vis_red}{red} boxes and \textbf{bold} fonts.}
\label{fig_instr_contrast}
\vspace{-0.2cm}
\end{figure*}

\subsection{Low-Resource Experiments}
We further conduct a low-resource experiment to verify how our RAM augmentation data can help mitigate the data scarcity issue. To this end, we only extract partial original data for the training of baseline~\cite{Chen2022ThinkGA}. 
Concretely, we randomly extract the data at different proportions and there may be large distributional discrepancies among extracted training subsets at different proportions.
For RAM, we add our rewritten data augmented from the corresponding partial original data. From 
Table~\ref{tab:low resource}
we can find that our RAM surpasses the baseline when extracting original data with different proportions. Especially, our RAM achieves comparable performance to the baseline which uses total original training data in 
Table~\ref{tab:training ablation} 
under the 60\% setting.
An interesting phenomenon we can also find in Table~\ref{tab:low resource} is that the baseline appears to generalize better with 40\% data compared to
60\%, which may be caused by large distributional discrepancies among extracted training subsets at different ratios. Moreover, it is also reasonable since it is common that the relationship between performance improvements and data scale is not linear, as also be demonstrated in Fig.~\ref{fig:mix training} for our augmentation data scale ablation.
These results further show the practicality of our RAM in real-world applications where the manually annotated data is usually limited.

\subsection{Visualization}
\noindent\textbf{Rewritten Observation-Instruction Examples.}
Fig.~\ref{fig_visual}
presents a visualization example of rewritten data by our RAM. From the rewritten object-enriched descriptions generated by the LLMs, we can find that diverse objects, e.g., {\it armchair} and {\it coffee table}, are indicated explicitly. Benefiting from it, the panorama generated by the T2IM contains these newly introduced objects with new spatial layouts, leading to great diversity enhancement of the environment. Moreover, we can observe that the extracted single-view images are also high-quality compared to the images taken by the camera in Matterport3D dataset. 
We can also find that the rewritten instruction successfully indicates modality-aligned objects, e.g., {\it hallway} and {\it plants} in the new trajectory, revealing the effectiveness of our observation-contrast instruction rewriting.


\noindent\textbf{Comparison between Speaker-based instructions and RAM instructions.}
We further give a visualization comparison of instructions generated by the Speaker model~\cite{li2024panogen} and our RAM in Fig.~\ref{fig_instr_contrast}, where we can find the instruction generated through the Speaker model is less informative, e.g. it contains repeated phrases ``walk around the table and chairs''. We can also observe that the instruction does not align well with the observations, e.g., the observations do not contain the {\it table} mentioned in the instruction, and the generated instruction changes the original actional information ``turn right'' to the wrong action ``turn left''. In contrast, the rewritten instruction by our RAM introduces reasonable actional representations as well as indicates multiple modality-aligned objects (such as {\it small kitchen} and {\it hammock}) that are different from those in the original instruction, which can effectively improve the cross-modal alignment ability of the agent.

\begin{figure}
\begin{centering}
\includegraphics[width=\linewidth]{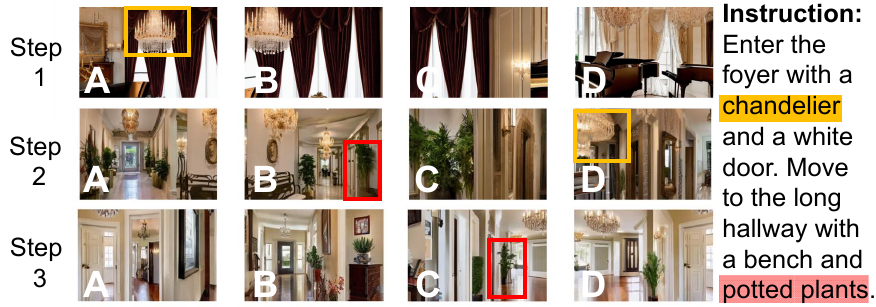}
\par\end{centering}
\caption{\label{fig:cross-step}Candidate observations for three consecutive steps. The cross-step semantically consisting objects are denoted in fonts/boxes in the same color.
}
\vspace{-0.4cm}

\end{figure}

\noindent\textbf{Cross-step Consistency.} We also give the visualization of candidate observations for multiple consecutive steps extracted from our rewritten panoramas in Fig.~\ref{fig:cross-step}, to study whether our rewritten observations can capture cross-step consistency. 
Since we use Text-to-Image Generation models (T2IMs) to generate rewritten panoramas for each step separately, there is no explicit constraint to ensure absolute cross-step observation overlap in our RAM. 
However, since our scene description rewriting mechanism can essentially generate or retain some consistent object categories for 
the same scene appearing in consecutive navigation timesteps, the generated panoramas can therefore have cross-step semantic consistency.
For example, 
the {\it potted plant} can be observed in both B in Step 2 and C in Step 3. Such cross-step semantic consistency can contribute to reasonable sequential action decisions, and our introduced rewritten data has also proved to be helpful for enhancing navigation performance from extensive experimental results. 


\section{Conclusion}
\label{conclusion}
In this work, we propose a novel \textbf{R}ewriting-driven \textbf{A}ug\textbf{M}entation (RAM) paradigm for the VLN task, 
where we create unseen observation-instruction pairs via performing object-enriched observation rewriting and observation-contrast instruction rewriting. 
Our RAM
fulfills simulator-free and labor-saving data augmentation, powered by delicate collaboration of various foundation models. We also introduce a mixing-then-focusing 
strategy accompanied by a random observation cropping scheme for boosting training with our rewritten data. 
Experimental results exhibit the impressive generalization ability of RAM 
on multiple popular VLN benchmarks. We believe our work can provide new insights for how to create and leverage augmentation data to address the critical data scarcity issue in embodied tasks.
In the future, we would like to introduce more effective mechanisms like parameter-efficient finetuning and feedback-based learning to further adapt foundation models for VLN data augmentation.

\bibliographystyle{IEEEtran}
\bibliography{IEEEabrv,tnnls}

\begin{IEEEbiography}[{\includegraphics[width=1in,height=1.25in,clip,keepaspectratio]{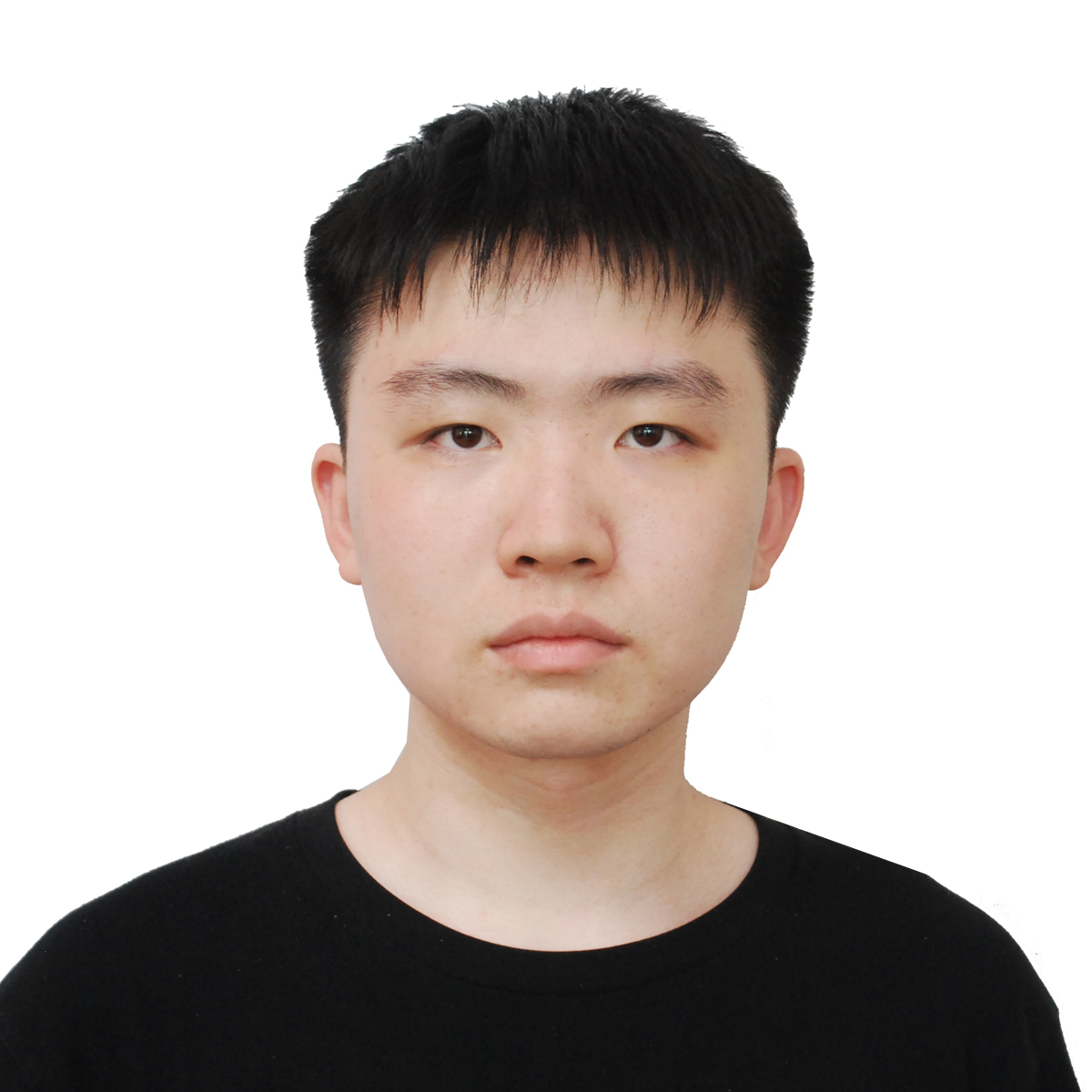}}]{Ziming Wei} received the B.E. degree in intelligence science and technology from Sun Yat-sen University in 2024. He is currently pursuing the M.S. degree with the school of intelligent systems engineering of Sun Yat-sen University. His current research interests include multi-modal understanding, learning and data generation, embodied AI and spatial intelligence.
\end{IEEEbiography}

\begin{IEEEbiography}[{\includegraphics[width=1in,height=1.25in,clip,keepaspectratio]{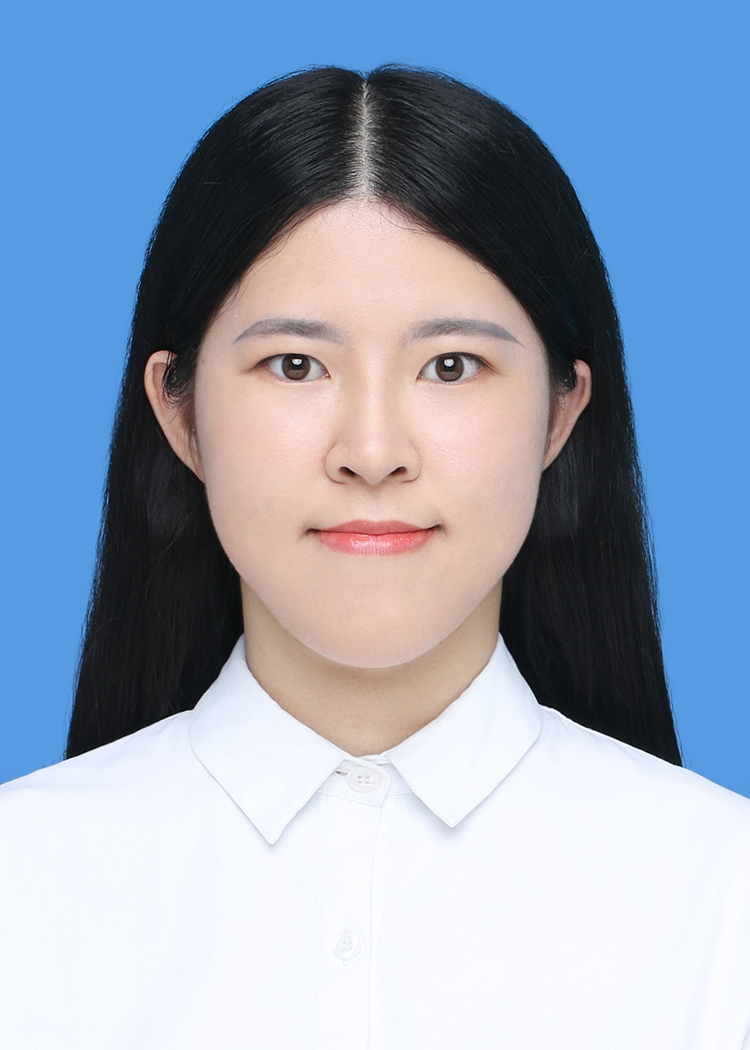}}]{Bingqian Lin} is currently a postdoc researcher at Shanghai Jiao Tong University, advised by Prof. Cewu Lu. She received her PhD degree from Sun Yat-sen University in 2024, advised by Prof. Xiaodan Liang and Prof. Liang Lin. She received the B.E. and the M.E. degree in Computer Science from University of Electronic Science and Technology of China and Xiamen University, in 2016 and 2019, respectively. Her research interests include vision-and-language understanding and embodied AI.
\end{IEEEbiography}

\begin{IEEEbiography}[{\includegraphics[width=1in,height=1.25in,clip,keepaspectratio]{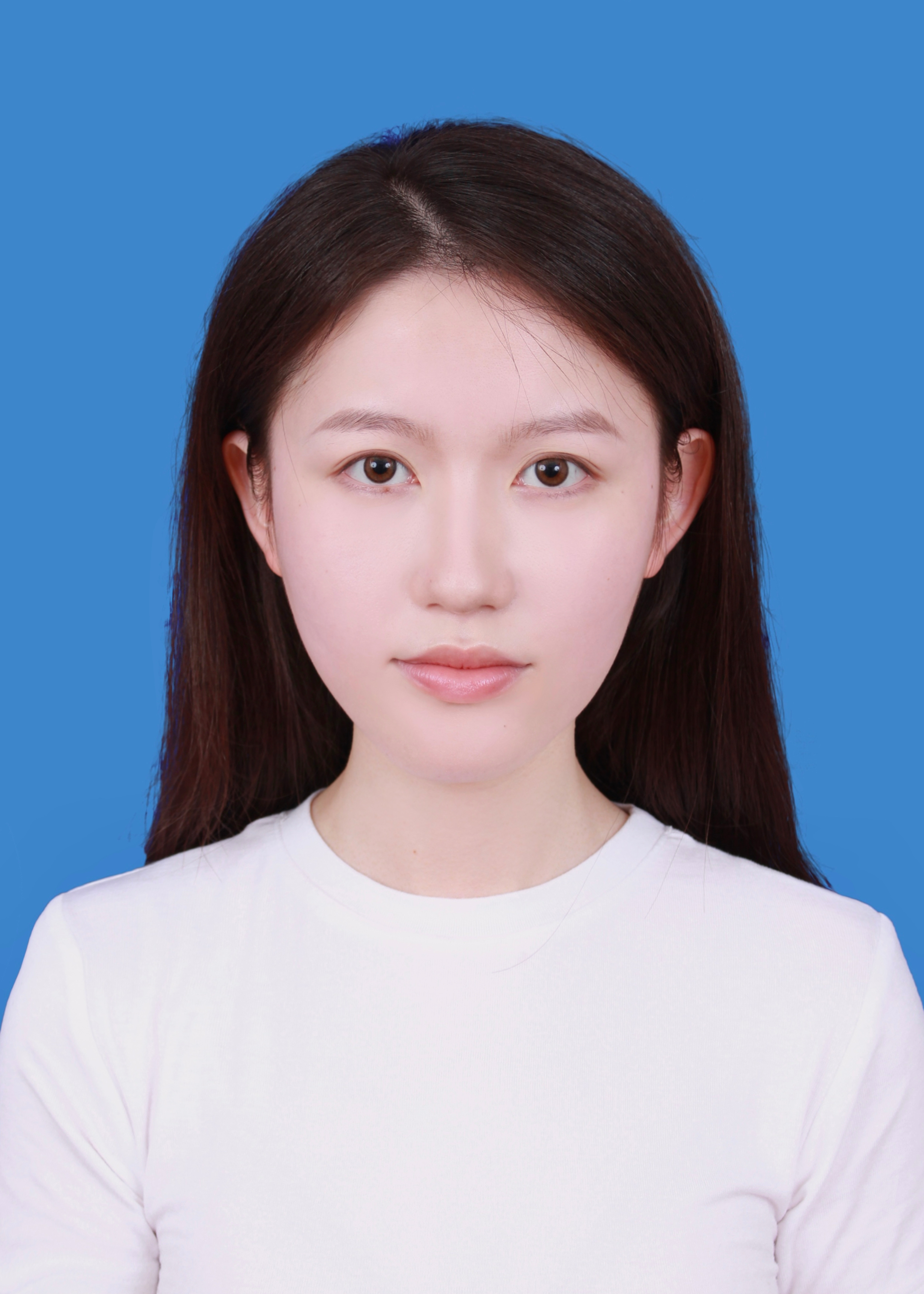}}]{Yunshuang Nie} received the B.E. degree in Sun Yat-sen University, Shenzhen, China, in 2023. She is currently working toward the M.E. in the school of intelligent systems engineering of Sun Yat-sen University. Her current research interests include vision-and-language understanding and embodied AI.
\end{IEEEbiography}

\begin{IEEEbiography}[{\includegraphics[width=1in,height=1.25in,clip,keepaspectratio]{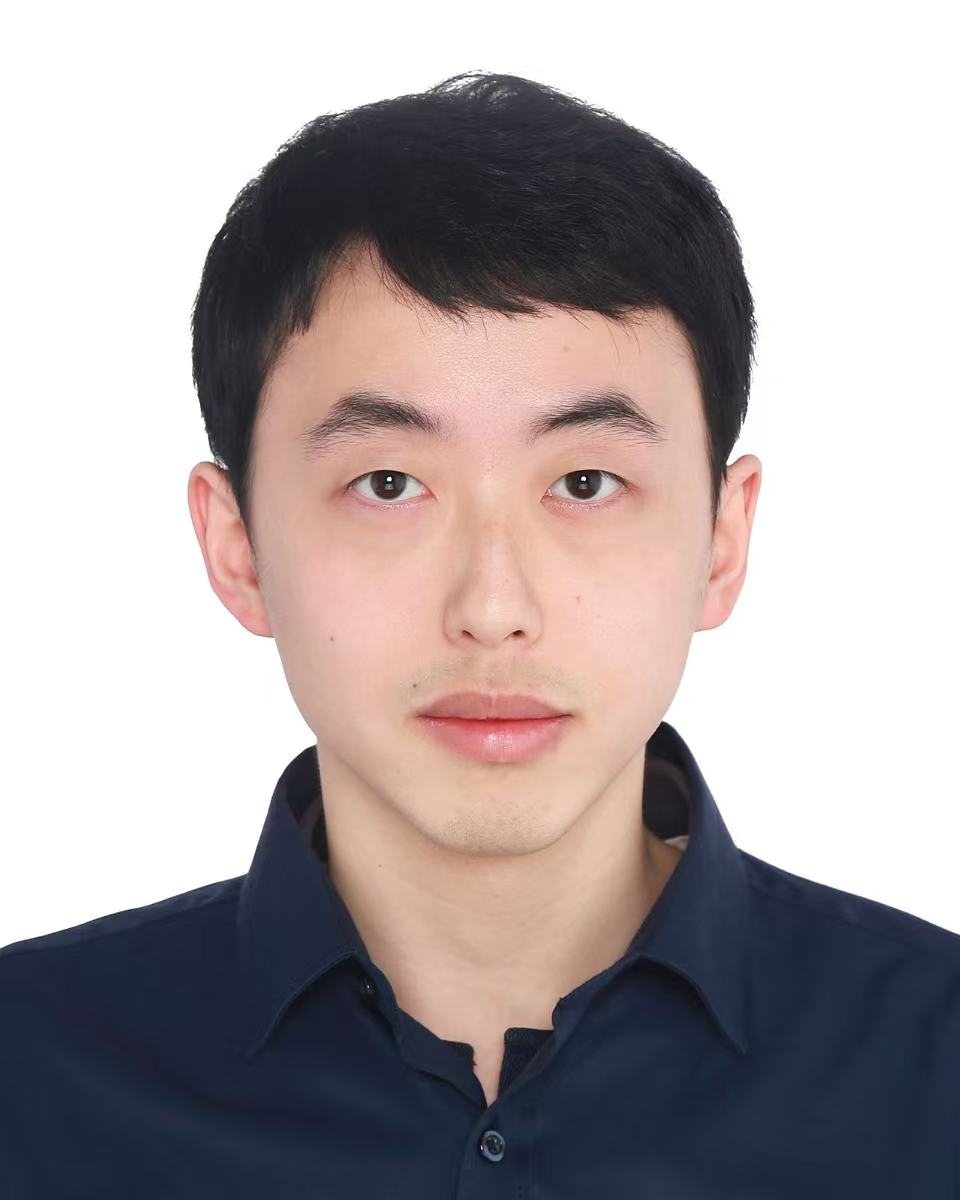}}]{Jiaqi Chen} received the B.E. and M.E. degrees from Xidian University and Sun Yat-sen University, in 2019 and 2021, respectively. He is currently working toward the Ph.D. degree in the Computer Science Department at The University of Hong Kong. His research interests include agent, multi-modal reasoning, and embodied AI.
\end{IEEEbiography}

\begin{IEEEbiography}
[{\includegraphics[width=1in,height=1.25in, clip,keepaspectratio]{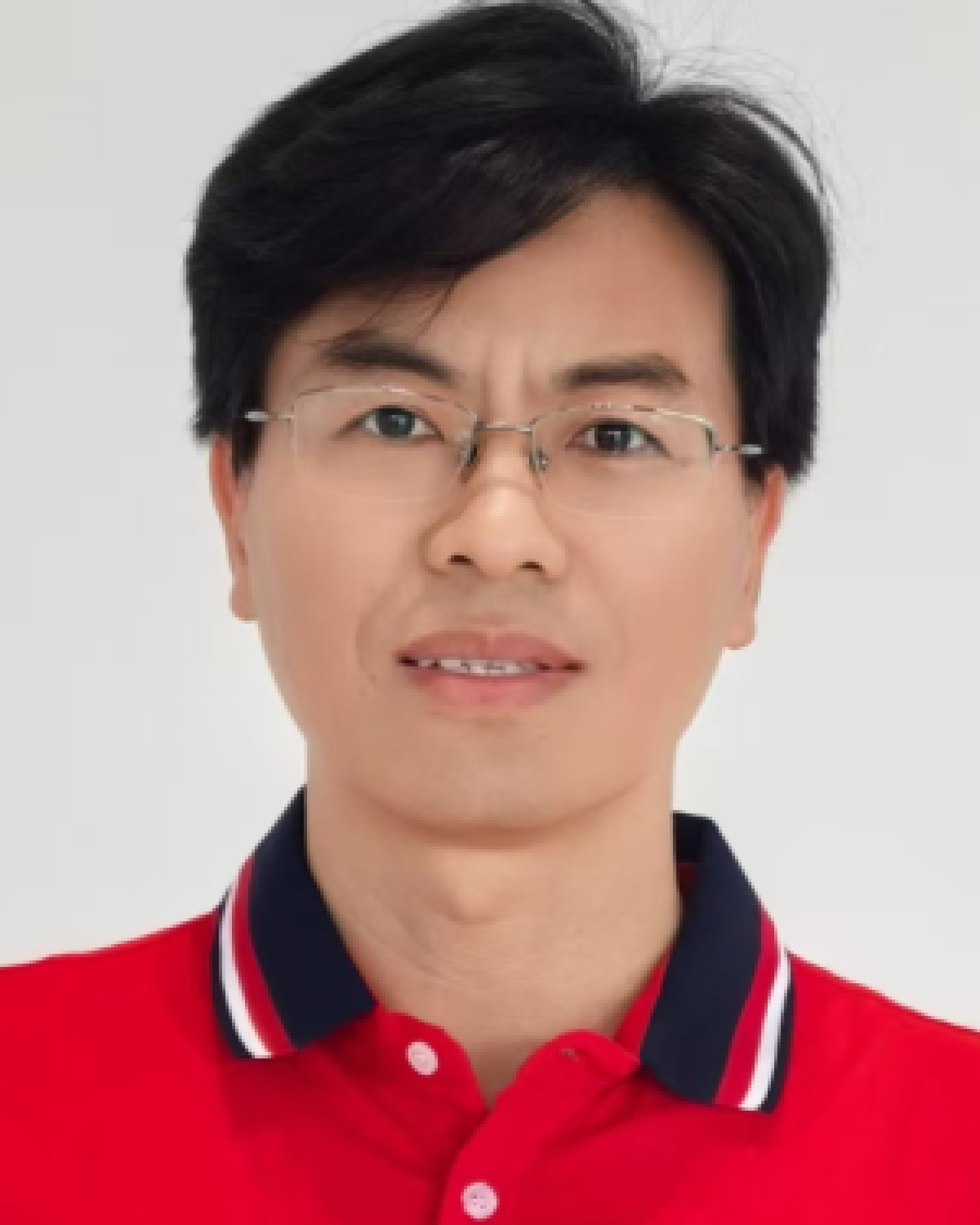}}]{Shikui Ma} received the M.S. degree from Northern Jiaotong University, Beijing, China, in 2003. He is currently the Chief Technology Officer of Hunan Artificial Intelligence and Robotics Institute Company Ltd. He has been serving as a Vice President of Dataa Robotics Company, Changsha, China, since 2015, where he is leading HARIX and AI Research and Development team to consistently
enhance their cloud robot brain HARIX platform, especially significantly improving the smart vision and motion capabilities of their robots, deep learning-based multi-modal fusion perception and advanced cognition, and deep reinforcement learning technologies.
\end{IEEEbiography}

\begin{IEEEbiography}
[{\includegraphics[width=1in,height=1.25in, clip,keepaspectratio]{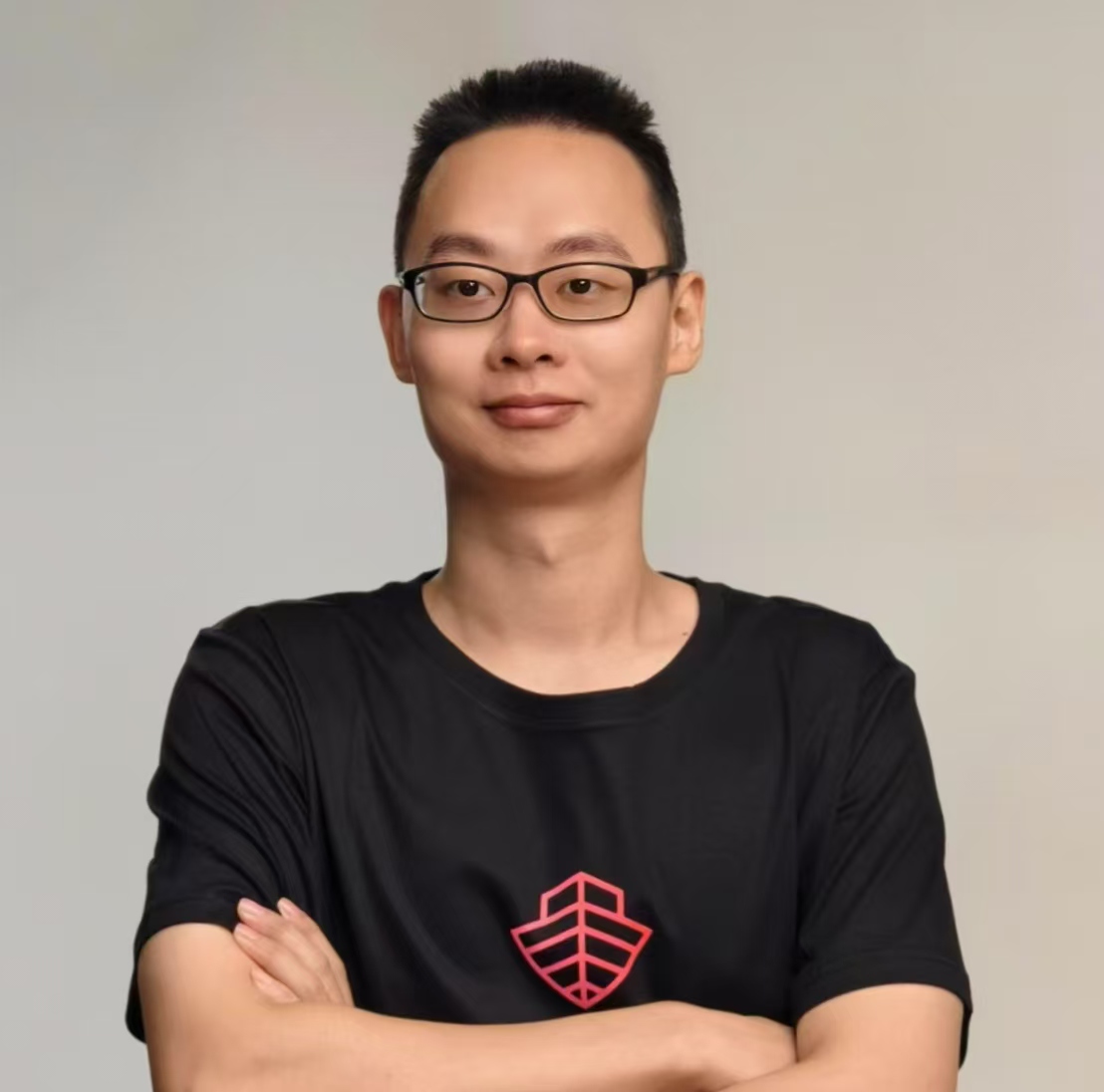}}]{Hang Xu} received the B.Sc. degree from Fudan University, Shanghai, China, in 2013, and the Ph.D. degree from The University of Hong Kong, Hong Kong, in 2018. He is currently a Senior CV Researcher at Huawei Noah’s Ark Lab, Shanghai. He has published over 100 papers at top AI conferences such as NeurIPS, CVPR, ICCV, and AAAI. His research interests include multimodal large language models, autonomous driving, object detection, and AutoML.
\end{IEEEbiography}

\begin{IEEEbiography}[{\includegraphics[width=1in,height=1.25in,clip,keepaspectratio]{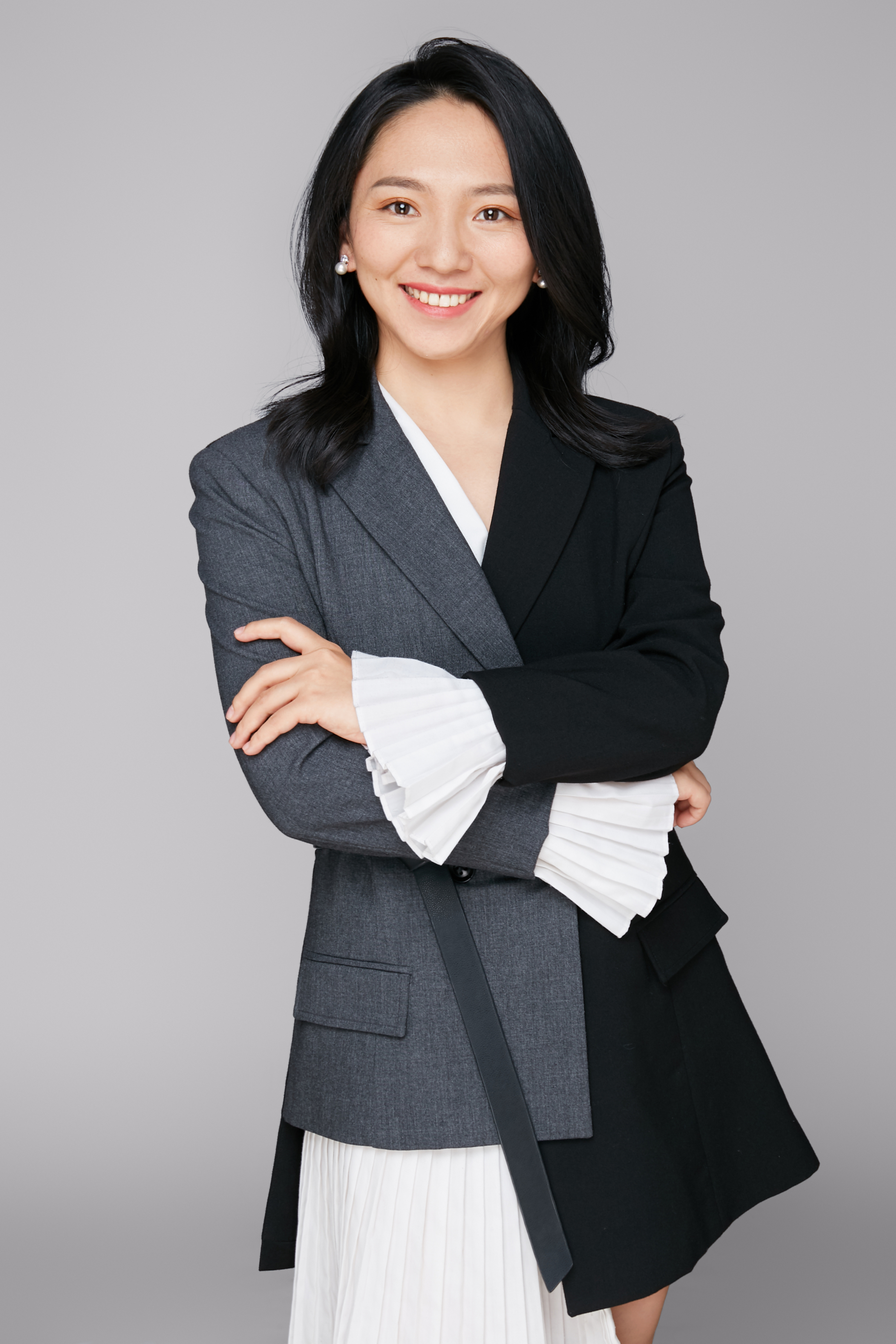}}]{Xiaodan Liang} is currently a Professor at Sun Yat-sen University.
She received the Ph.D. degree from Sun Yat-sen University in 2016,
advised by Liang Lin.
She was a Post-Doctoral Researcher with the
Machine Learning Department, Carnegie Mellon
University, Pittsburgh, PA, USA, from 2016 to 2018,
working with Prof. Eric Xing. She has published several cutting-edge projects
on human-related analysis, including human parsing, pedestrian detection and instance segmentation,
2D/3D human pose estimation, and activity recognition.

\end{IEEEbiography}

\clearpage
\begin{center}
    \LARGE\textbf{Supplementary Material}
\end{center}
\setcounter{section}{0}  
\section{Method Details}
To improve readability and clarity, we provide pseudocodes for our mixing-then-focusing strategy and the data augmentation pipeline, presented in Algorithm \ref{alg_mftm} and Algorithm \ref{alg_dap} on the last page, respectively.

\section{Implementation Details}

\subsection{Scale of augmented data}
For each dataset (R2R, REVERIE, and R4R), the augmented data are three times larger than the original dataset, shown in Table~\ref{tab_data_amount}, which is significantly smaller than other methods such as AutoVLN~\cite{chen2022learning} and ScaleVLN~\cite{wang2023scaling}. All our augmentation data are utilized in both pretraining and finetuning phases.

\begin{table}[h]
	\fontsize{18}{18}\selectfont
	\vspace{-0.2cm}
	\caption{Scale of augmented data for our RAM and ScaleVLN~\cite{wang2023scaling}.}	
	\label{tab_data_amount}
 \vspace{-0.4cm}
	\centering
	\resizebox{\linewidth}{!}{
	{\renewcommand{\arraystretch}{1.3}\begin{tabular}{c||c c|c}
			 \specialrule{.1em}{.05em}{.05em}
		 
         Dataset&R2R \& R4R (ours)&REVERIE (ours)&R2R (ScaleVLN)
    
    \cr\hline
        Amount of Augmented Data&14,025&12,450&4,941,710\\
        
        
          \specialrule{.1em}{.05em}{.05em}
		\end{tabular}}}
\vspace{-0.6cm} 
\end{table}

\subsection{Choices of foundation models and implementation details}
For generating observations, the reason for selecting Multidiffusion~\cite{bar2023multidiffusion} as the text-to-image model used in our pipeline has been indicated in our original manuscript, i.e., the Multidiffusion model is more suitable for generating panoramas. In the original Multidiffusion~\cite{bar2023multidiffusion} paper, the authors also demonstrated its superior capability of panorama generation compared to Stable Diffusion.

For generating scene descriptions, we have compared the descriptions obtained through different popular VLMs, including BLIP~\cite{li2022blip}, Tag2Text~\cite{huang2023tag2text}, and LLaVA~\cite{liu2023llava}, and finally chose Tag2Text~\cite{huang2023tag2text} since it tends to capture informative and salient objects in the observation image. We also put a comparison visualization example here in Fig.~\ref{fig:captions_copy1}. The results in Fig.~\ref{fig:captions_copy1} reveal that BLIP tends to miss critical objects in the scene and LLaVA often generates redundant or irrelevant information for image generation. In contrast, Tag2Text can accurately identify noteworthy objects and generate informative captions.

\begin{table*}[t]
	\fontsize{18}{18}\selectfont
	\caption{Details on the time, computational cost, and resources required for generating the augmented data by our RAM.}	
	\label{tab_cost}
	\centering
	\resizebox{0.8\linewidth}{!}{
	{\renewcommand{\arraystretch}{1.3}\begin{tabular}{c||c|c||c|c|c|c}
			 \specialrule{.1em}{.05em}{.05em}
		 
         No.&Data Collection Process&Foundation Models&API Querying&Money Cost&GPUs&Time Consumption
    
    \cr\hline
        1&Observation Descriptions&VLM&\ding{55}&-&\ding{51}&$<$ 1 hours\\
        2&Rewritten Observation Descriptions&LLM&\ding{51}&10 dollars&\ding{55}&$<$ 30 minutes\\
        3&Augmented Observations&T2I Model&\ding{55}&-&\ding{51}&$\sim$ 30 hours\\
        4&Rewritten Instructions&LLM&\ding{51}&10 dollars&\ding{55}&$<$ 30 minutes\\
        
        
          \specialrule{.1em}{.05em}{.05em}
		\end{tabular}}}

\end{table*}

\definecolor{moti_red}{RGB}{192,0,0}
\definecolor{moti_blue}{RGB}{0,153,255}
\begin{figure*}[t]
\vspace{-0.2cm}
\begin{centering}
\includegraphics[width=0.7\linewidth]{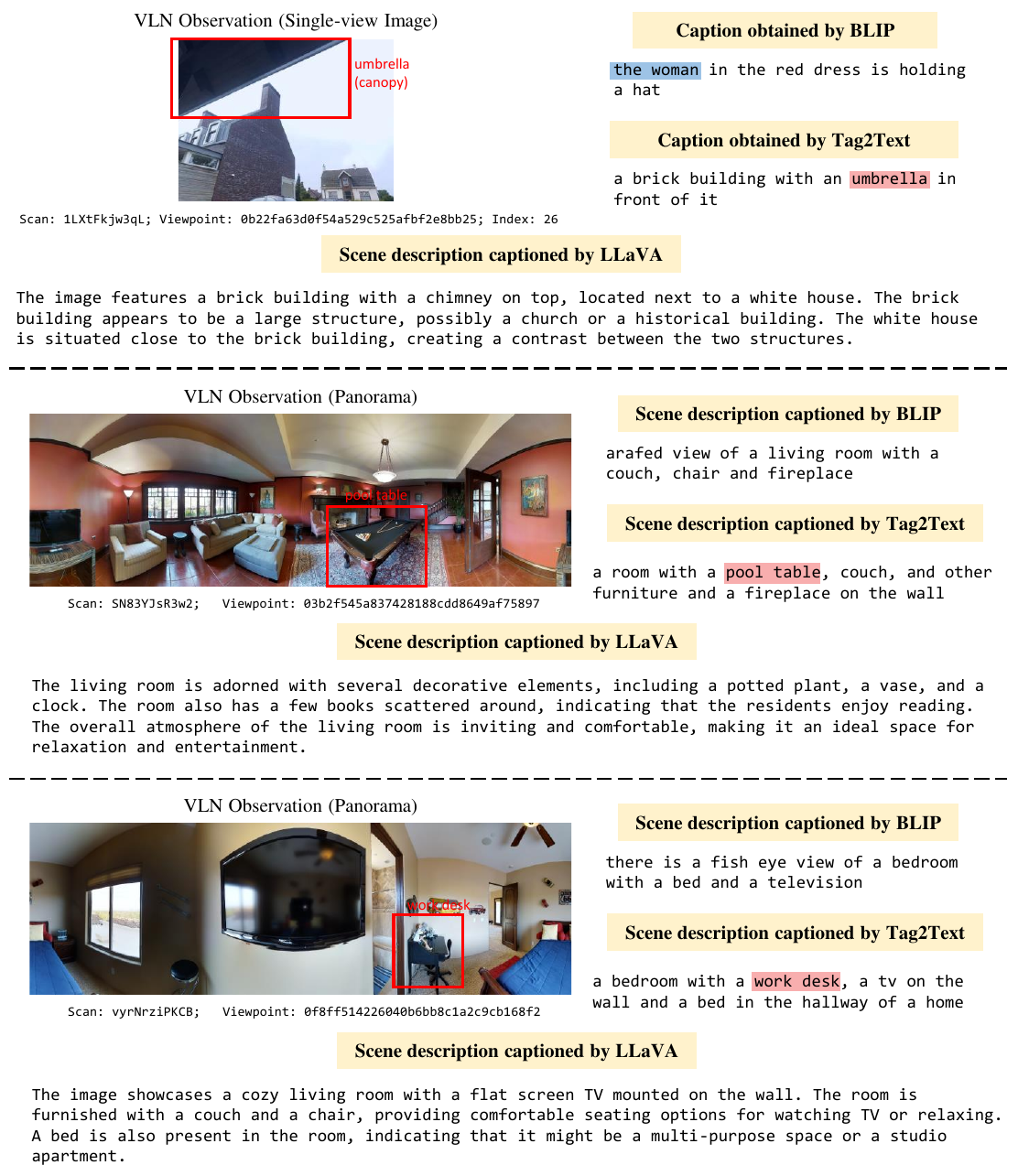}
\par\end{centering}

\caption{Captions of single-view images and scene descriptions of panoramas obtained by different VLMs~\cite{li2022blip, huang2023tag2text}. Scene objects are denoted in \textcolor{moti_red}{red} fonts/boxes, and the incorrect noun in the caption is denoted in \textcolor{moti_blue}{blue} box.}
\label{fig:captions_copy1}
\vspace{-0.4cm}
\end{figure*}

Regarding the choice of LLM, we selected \textit{gpt-3.5-turbo-1106} by considering its low cost and satisfactory performance. While more advanced and expensive LLMs, such as GPT-4, might generate better scene descriptions, it may be constrained by the marginal benefit of performance improvements.

The VLM, LLM, and T2IM used in our data augmentation pipeline are all employed in an off-the-shelf manner. For the VLM (i.e., Tag2Text~\cite{huang2023tag2text}), we keep the original parameter setting (e.g. \textit{threshold for tagging} and \textit{input image size}) for obtaining scene descriptions. For the LLM (i.e., ChatGPT~\cite{openai}), we set the parameter \textit{temperature} to 0.8 and \textit{presence\_penalty} to 0. For the T2IM (i.e., MultiDiffusion~\cite{bar2023multidiffusion}), we
fix the parameter \textit{num\_inference\_steps} of the MultiDiffusion model to 30 for generating panoramas with high quality and resolution in a tolerable time consumption.

\subsection{Details on the time, computational cost and resources}
Details on the time, computational cost, and resources required for generating the augmented data by our RAM are presented in Table~\ref{tab_cost}. Concretely, we utilized VLM and T2I models to obtain observation descriptions and augmented observations on 8 NVIDIA RTX 3090 GPUs by splitting data into multiple batches in 1 hour and 30 hours, respectively. 
Rewritten observation descriptions and rewritten instructions were collected by querying LLMs within 30 minutes. Overall, the entire data augmentation pipeline requires less than $\sim$2 days and relatively low computational resources.


\begin{table}[h]
	\fontsize{12}{12}\selectfont
\centering
\caption{Comparison results with the baseline HAMT~\cite{Chen2021HistoryAM} on R2R dataset. 
\dag: using CLIP ViT-B/16 as image features.}
	\label{tab1:new_baseline_r2r}
	\resizebox{\linewidth}{!}{
	{\renewcommand{\arraystretch}{1.2}
		\begin{tabular}{c||c|c|c|c|c|c|c|c}

			\specialrule{.1em}{.05em}{.05em}
			\multirow{2}{*}{Method}&\multicolumn{4}{c|}{Val Seen }&\multicolumn{4}{c}{Val Unseen}\cr\cline{2-9}
			&TL&NE $\downarrow$&SR $\uparrow$&SPL $\uparrow$&TL&NE $\downarrow$&SR $\uparrow$&SPL $\uparrow$\cr
			\hline

    HAMT$^{\dag}$~\cite{Chen2021HistoryAM} (baseline)&11.80&3.08&70.52&66.80&12.06&4.09&62.03&56.62\\
    HAMT+RAM$^{\dag}$(ours)&\textbf{10.94}&\textbf{2.82}&\textbf{73.95}&\textbf{71.11}&\textbf{11.45}&\textbf{3.91}&\textbf{63.26}&\textbf{59.27}\\

 \specialrule{.1em}{.05em}{.05em}

		\end{tabular}}}
	\vspace{-0.5cm}
\end{table}

\section{More Experiments and Visualization}
\subsection{Performance comparison with baseline HAMT on R2R}
We have incorporated a new baseline, HAMT~\cite{Chen2021HistoryAM}, along with corresponding experiments on the R2R dataset, as presented in Table \ref{tab1:new_baseline_r2r} here. From Table \ref{tab1:new_baseline_r2r}, we can observe that our RAM outperforms the baseline HAMT in both Val Seen and Val Unseen, e.g., it brings $\sim$3\% improvements in SPL in Val Unseen. The results on different baselines demonstrate the effectiveness and generalization ability of our proposed method.

\begin{figure*}[h]
\begin{centering}
\includegraphics[width=0.85\linewidth]{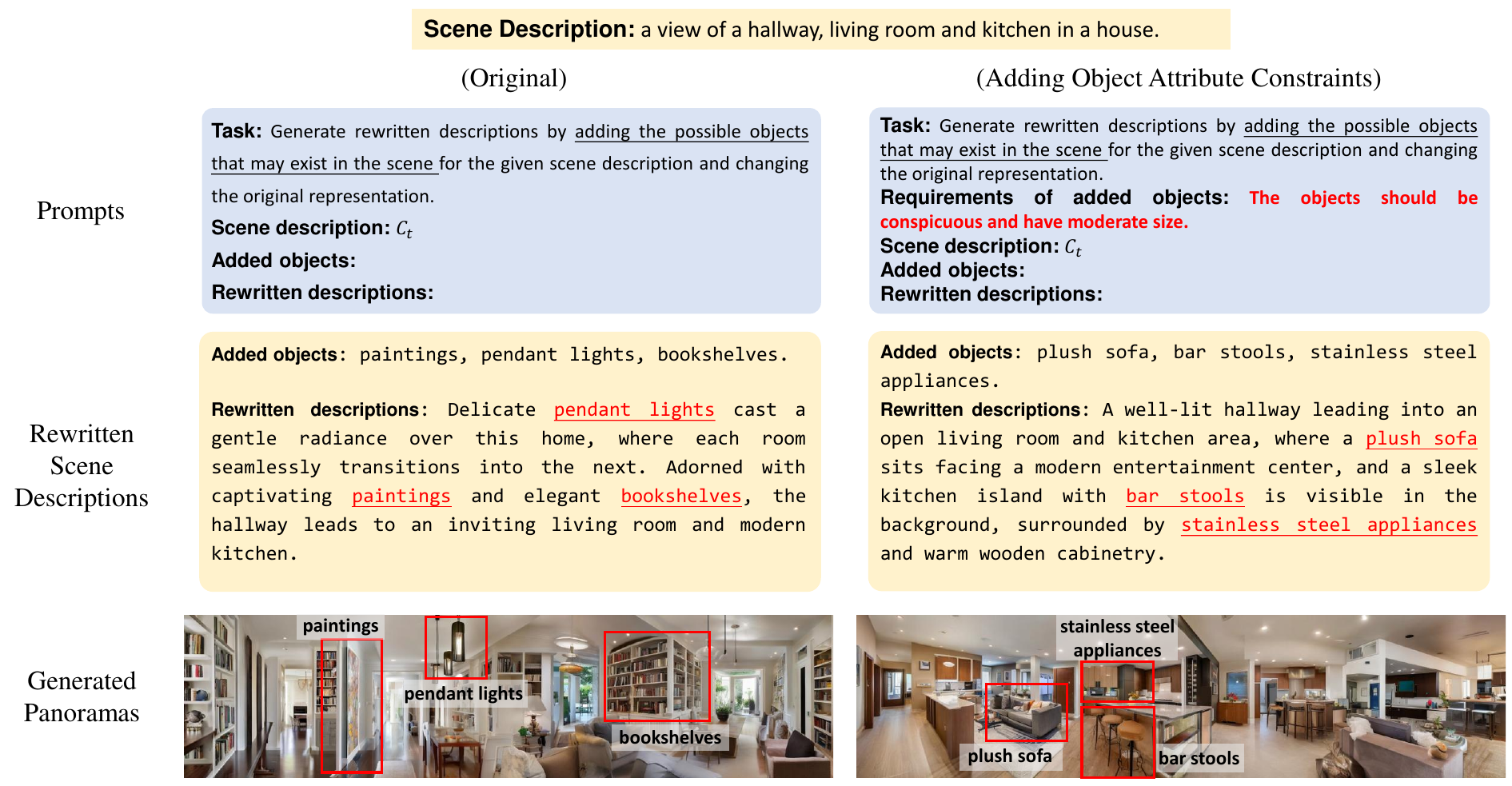}
\par\end{centering}

\caption{Rewritten descriptions with different prompts and panoramas generated by T2I model with the corresponding rewritten descriptions.
On the right side, we utilize the prompt adding object attribute constraints to generate rewritten descriptions. 
Added objects in rewritten scene descriptions and generated panoramas are denoted in \textcolor{red}{red} boxes and \textbf{bold} fonts.}
\label{fig:rd}
\vspace{-0.4cm}
\end{figure*}

\subsection{Comparison with ScaleVLN on the same data scale}
Table~\ref{tab:scalevln} shows a new ablation experiment with CLIP VIT-B/16 visual features. Considering that  acquiring data at the scale of ScaleVLN~\cite{wang2023scaling} is highly labor-intensive and poses significant storage challenges, we conduct an alternative experimental comparison between ScaleVLN~\cite{wang2023scaling} and our RAM. 
Specifically, we train DUET~\cite{Chen2022ThinkGA} agent using the ScaleVLN data with the same amount of augmented data as our RAM, i.e., we randomly sample 14,025 trajectory-instruction pairs (3$\times$ the data amount of the original R2R data) from the ScaleVLN~\cite{wang2023scaling} dataset in R2R dataset\footnote{The CLIP VIT-B/16 visual features are provided by the ScaleVLN GitHub repository, https://github.com/wz0919/ScaleVLN.}. The results are presented in Table~\ref{tab:scalevln}. From Table~\ref{tab:scalevln}, we can observe that the ``DUET+ScaleVLN(subset)'' setting surpasses the baseline, yet its performance is inferior to that of our RAM. Moreover, this setting's performance on the Val Seen drops by $\sim$6\% in SR and $\sim$8\% in SPL compared to the baseline, which may be attributed to the introduction of a large amount of unfamiliar real-world simulator data causing the agent to lose its generalization capacity in the original simulation environment. Additionally, we can see that our RAM outperforms ScaleVLN in Val Unseen with the same data scale, which further validates the high quality of our augmented data and the effectiveness of our proposed mixed training mechanism.

\begin{table}[t]
	\fontsize{12}{12}\selectfont

\caption{Comparison with baseline method~\cite{Chen2022ThinkGA} and ScaleVLN~\cite{wang2023scaling} with the same scale as ours (3$\times$ larger than the R2R dataset). \dag: using CLIP ViT-B/16 as visual features.}
        \centering
	\label{tab:scalevln}
	\resizebox{\linewidth}{!}{
	{\renewcommand{\arraystretch}{1.2}
		\begin{tabular}{c||c|c|c|c|c|c|c|c}

			\specialrule{.1em}{.05em}{.05em}
			\multirow{2}{*}{Method}&\multicolumn{4}{c|}{Val Seen }&\multicolumn{4}{c}{Val Unseen}\cr\cline{2-9}
			&TL&NE $\downarrow$&SR $\uparrow$&SPL $\uparrow$&TL&NE $\downarrow$&SR $\uparrow$&SPL $\uparrow$\cr
            
    \hline
    
    DUET$^{\dag}$~\cite{Chen2022ThinkGA} (baseline)&13.86&2.12&80.80&73.64&16.22&3.06&72.37&58.75\\

    DUET+ScaleVLN(subset)$^{\dag}$&13.84&2.75&74.14&65.62&13.67&3.11&73.01&62.60\\
    DUET+RAM$^{\dag}$(ours)&\textbf{11.50}&\textbf{1.95}&\textbf{82.17}&\textbf{77.70}&\textbf{13.33}&\textbf{2.96}&\textbf{73.65}&\textbf{63.13}\\

 \specialrule{.1em}{.05em}{.05em}

		\end{tabular}}}
	\vspace{-0.3cm}
\end{table}

\subsection{Comparison with object deletion}
To realize object deletion, we mask some specific objects in the observations following EnvEdit~\cite{Li2022EnvEditEE} and remove the corresponding objects from the rewritten instructions. We present the results here in Table \ref{tab:deletion}, where we can find that   object deletion can also improve performance, indicating that this approach is another effective way to generate meaningful augmentation data. It provides a valuable reference for designing future data augmentation approaches. 
Furthermore, the results in Table~\ref{tab:deletion} show that our RAM outperforms the object deletion, suggesting that introducing a greater variety of objects can potentially enhance observation diversity and facilitate the agent's learning of cross-modal alignment, thereby improving navigation performance. 

\begin{table}[t]
	\fontsize{18}{18}\selectfont
	\caption{Ablation results on object addition and object deletion. We conduct this experiment using CLIP ViT-B/16 as visual features.}	
	\label{tab:deletion}
	\centering
			\resizebox{0.45\linewidth}{!}{
	{\renewcommand{\arraystretch}{1.2}\begin{tabular}{c||c|c}
			 \specialrule{.1em}{.05em}{.05em}
		 
         \multirow{2}{*}{Settings}&\multicolumn{2}{c}{Val Unseen}\cr\cline{2-3}
    &SR$\uparrow$&SPL$\uparrow$
    
    \cr\hline
        baseline&72.37&58.75\\
        object deletion&72.46&61.17\\
        RAM(ours)&\textbf{73.65}&\textbf{63.13}\\
          
        
          \specialrule{.1em}{.05em}{.05em}
		\end{tabular}}}
\vspace{-0.2cm}
\end{table}

\subsection{Controlled object generation}
To investigate the potential of our RAM in controlled object generation for creating specific VLN scenarios, we incorporate the object attribute constraints on the LLM prompt presented in Fig.~\ref{fig:rd}. From Fig.~\ref{fig:rd}, we can observe that by adding the constraints, the LLM can successfully generate objects with desired styles, e.g., the ``bar stools'' and ``plush sofa'' which have a moderate size. And the T2IM also successfully generates the desired objects in the panoramas. These results reveal the potential of extending our method to generate VLN scenarios with specific styles.

\newpage
\vspace{-0.2cm}
\begin{algorithm*}[t]
	\caption{Mixing-then-Focusing Training Mechanism} 
	\label{alg_mftm} 
        \setstretch{1.3}
        \SetKwInOut{Input}{Input}
        \SetKwInOut{Output}{Output}
		\Input{original real-world dataset $\mathcal{D}_{train}=\left\{\left\{O_{t}\right\}^{T}_{t=1}, I\right\}$, our rewritten dataset $\mathcal{D}_{train}=\left\{\left\{O^{r}_{t}\right\}^{T}_{t=1}, I^{r}\right\}$, navigation agent $E^{n}$}
        \Output{trained agent $E^{n}_{s2}$}
	\begin{algorithmic}[1]
        \STATE \textit{for iteration in} \textbf{Stage 1} \textit{do:}
        \STATE\hspace{\algorithmicindent}Use random observation cropping scheme (denoted as $\mathrm{RC}(\cdot)$) to augment $\left\{O^{r}_{t}\right\}^{T}_{t=1}$, i.e.,  $\left\{\mathcal{O}^{r}_{t}\right\}^{T}_{t=1} = \mathrm{RC}\left(\left\{O^{r}_{t}\right\}^{T}_{t=1}\right)$;
        \STATE\hspace{\algorithmicindent}Train $E^{n}$ with $\mathcal{D}_{train}=\left\{\left\{O_{t}\right\}^{T}_{t=1}, I\right\}$ and $\mathcal{D}^{'}_{train}=\left\{\left\{\mathcal{O}^{r}_{t}\right\}^{T}_{t=1}, I^{r}\right\}$;
        \STATE\hspace{\algorithmicindent}Optimize $E^{n}$ based on loss $\mathcal{L}_{s1}$ and get trained agent $E^{n}_{s1}$.
        \STATE \textit{end}

        \hspace{\algorithmicindent}
        
        \STATE \textit{for iteration in} \textbf{Stage 2} \textit{do:}
        \STATE\hspace{\algorithmicindent}Train $E^{n}_{s1}$ with $\mathcal{D}_{train}=\left\{\left\{O_{t}\right\}^{T}_{t=1}, I\right\}$;
        \STATE\hspace{\algorithmicindent}Optimize $E^{n}_{s1}$ based on loss $\mathcal{L}_{s2}$ and get trained agent $E^{n}_{s2}$.
        \STATE \textit{end}
	\end{algorithmic} 
\end{algorithm*}
\vspace{-0.5cm}

\begin{algorithm*}[t]
	\caption{Data Augmentation Pipeline} 
	\label{alg_dap} 
        \setstretch{1.4}
        \SetKwInOut{Input}{Input}
        \SetKwInOut{Output}{Output}
		\Input{original panoramic observation $O_{t}$, original instruction $I$, scene description rewriting prompt $P^{c}$, instruction rewriting prompt $P^{i}$}
        \Output{rewritten observations $O^{r}_{t}$, rewritten instruction $I^{r}$.}
        \begin{algorithmic}[1]
        \STATE \textit{for iteration in} \textbf{Object-Enriched Observation Rewriting} \textit{do:}
        \STATE\hspace{\algorithmicindent}\textbf{(1) Object-Enriched Scene Description Rewriting:}
        \STATE \hspace{\algorithmicindent}\hspace{\algorithmicindent}Use the Vision-Language Model (VLM) to collect panoramic observation descriptions, i.e.,  $C_{t}=\mathrm{VLM}\left(O_{t}\right)$
        \STATE\hspace{\algorithmicindent}\hspace{\algorithmicindent}Obtain the rewritten object-enriched scene descriptions $C^{r}_{t}$ accompanied by added objects $\{B_{t,n}\}_{n=1}^{N}$ with $P^{c}$ and $C_{t}$, i.e.,  $C^{r}_{t}, \left\{B_{t,n}\right\}_{n=1}^{N}=\mathrm{LLM}\left(C_{t}, P^{c}\right)$
        
        \STATE\hspace{\algorithmicindent}\textbf{(2) Panorama-to-View Observation Generation:}
        \STATE \hspace{\algorithmicindent}\hspace{\algorithmicindent}Obtain rewritten observations $O^{r}_{t}$ by feeding $C^{r}_{t}$ into the panoramic T2IM model, i.e., $O^{r}_{t} = \mathrm{T2IM}\left(C^{r}_{t}\right)$
        \STATE \hspace{\algorithmicindent}\hspace{\algorithmicindent}Use $\mathrm{Equirec2Perspec}$ algorithm to get single-view images, i.e., $\{O^{r}_{t,k}\}_{k=1}^{K} = \mathrm{Equirec2Perspec}\left(J_{inv}, R, O^{r}_{t}\right)$
        \STATE \textit{end}

        \hspace{\algorithmicindent}
        \STATE \textit{for iteration in} \textbf{Observation-Contrast Instruction Rewriting} \textit{do:}
        \STATE \hspace{\algorithmicindent}\textbf{(1) Sequential Landmark Grounding:}
        \STATE \hspace{\algorithmicindent}\hspace{\algorithmicindent}Use an LLM to extract the sequential landmarks $U=\{U_{k}\}_{k=1}^{M}$ from the original instruction $I$
        \STATE \hspace{\algorithmicindent}\hspace{\algorithmicindent}
        Extract the ground-truth action (observation) $G_{t}$ at timestep $t$  from the original observation $O_{t}$ \STATE \hspace{\algorithmicindent}\hspace{\algorithmicindent}Employ CLIP~\cite{radford2021learning} to find the matched landmark $U_{t}$ for each $G_{t}$ by taking the landmark $U_{k}$ which has maximum similarity with $G_{t}$
        \STATE \hspace{\algorithmicindent}\textbf{(2) New Observation Description Collection:}
        \STATE \hspace{\algorithmicindent}\hspace{\algorithmicindent}
        Extract the ground-truth action (observation) $G'_{t}$ which has
the same position as $G_{t}$ from the rewritten observations $O^{r}_{t}$
        \STATE \hspace{\algorithmicindent}\hspace{\algorithmicindent}Adopt the VLM to generate the description $C'_{t}$ for the ground-truth action (observation) $G'_{t}$, i.e.,  $C'_{t}=\mathrm{VLM}\left(G'_{t}\right)$
        \STATE \hspace{\algorithmicindent}\textbf{(3) Instruction Rewriting by Observation Contrast:}
        \STATE \hspace{\algorithmicindent}\hspace{\algorithmicindent}Acquire the rewritten instruction $I^{r}$ by querying the LLM, i.e., $I^{r} = \mathrm{LLM}(\{U_{t}\}_{t=1}^{T}, \{C'_{t}\}_{t=1}^{T},  I, P^{i})$
        \STATE \textit{end}
	\end{algorithmic} 
\end{algorithm*}

\end{document}